\documentclass[10pt,twocolumn,letterpaper]{article}

\usepackage{cvpr}
\usepackage{times}
\usepackage{epsfig}
\usepackage{graphicx}
\usepackage{amsmath}
\usepackage{amssymb}

\usepackage[lined,algonl,boxed]{algorithm2e}
\usepackage{color} 

\def\bx{{\bf x}}

\def\bUpsilon{\boldsymbol{\Upsilon}}

\cvprfinalcopy 

\def\cvprPaperID{757} 

\ifcvprfinal\fi
\begin{document}

\title{3D planar patch extraction from stereo using probabilistic region growing}

\author{Vasileios Zografos\\
Linkoping University\\
Sweden\\
{\tt\small zografos@isy.liu.se}}

\maketitle

\begin{abstract}
   This article presents a novel 3D planar patch extraction method using a probabilistic region growing algorithm. Our method works by simultaneously initiating multiple planar patches from seed points, the latter determined by an intensity-based 2D segmentation algorithm in the stereo-pair images. The patches are grown incrementally and in parallel as 3D scene points are considered for membership, using a probabilistic distance likelihood measure. In addition, we have incorporated prior information based on the noise model in the 2D images and the scene configuration but also include the intensity information resulting from the initial segmentation.
   This method works well across many different data-sets, involving real and synthetic examples of both regularly and non-regularly sampled data, and is fast enough that may be used for robot navigation tasks of path detection and obstacle avoidance. 
\end{abstract}

\section{Introduction}
Extraction of 3D planar patches can be very useful in robotic navigation applications where 3D structure data is available and we wish to determine the approximate main navigable areas and obstacles along the way. Additionally, such an approach may be employed on: building/scene reconstruction from sensory point data; data simplification and compression, since data may be represented with a simple parametric model; de-noising of a 3D scene, since the model estimated from several measurements will contain a reduced amount of bias and uncertainty; perceptual organisation and 3D segmentation, texture extraction and dense resampling of data from the fitted parametric models.

We present an incremental method for extracting such planar areas in 3D scenes using a region growing algorithm, whereby planes are initially fitted to a selection of 3D points and are incrementally grown as new points are added. The parametric plane models are gradually refined as a function of the influence of each 3D point, here determined probabilistically, resulting in a geometrically consistent and perceptually meaningful extraction. 

We consider  a set of 3D points $\bUpsilon$ and projection mappings $\bx$, $\bx'$,
which define the transformation from world coordinates to image $I$,
$I'$ coordinates. In order to define these mappings, it is assumed
that the camera positions and calibration are given. The proposed algorithm  contains three distinct steps: patch \emph{seeding}, in which small clusters of points are used to initialise the patch growing process. The centres of these clusters are determined by the consistent segmentation of both images $I$,
$I'$ ; patch \emph{growing}, where these initial patches are grown by adding new 3D points using a probabilistic criterion and patch \emph{refinement} whereby patches are merged or simply discarded using a number of quantitative measures.

The novelty of our work includes the solution of the incremental patch growth task from a probabilistic viewpoint. We formulate a Bayesian inference model and incorporate subjective prior probabilities to combine our knowledge on the specifics of the 3D scene configuration, stereo-pair camera projection and noise model, with the purely geometrical observations. In addition, we make use of a 2D segmentation pre-step, consistent in both camera views, as a means to initialise our patches in perceptually meaningful regions and also in order to incorporate pixel intensity information in our prior model.

\section{Background\label{sec:background}}
Plane extraction from 3D data is an area that has been explored extensively in the past (see ~\cite{Hoo1996} for an overview of the area).
The recent, available methods fall into two main areas: iterative extraction, which include sequential growing algorithms and split and merge methods; and point clustering approaches that attempt to classify as many points as possible using a planar constraint criterion usually with the aid of some optimised sampling strategy. We focus here on the former area, which is the most closely related to our approach. Good examples are the work by~\cite{BalTra2000} where they use a dual, iterative method to establish the plane parameters from the translational component of the camera motion. In \cite{LeeSch2002} the authors propose a bottom-up method for organisation of points into perceptually meaningful areas with an emphasis on planar regions. This is the method most similar to ours in the sense that it seeds and grows planes, but only based on the geometry and without fusing scene information in a probabilistic manner. Their method focuses more on the higher, semantic levels  whereas we focus on the lower, signal levels.

A split and merge method based on purely the plane equation is proposed by~\cite{TseWan2005} which continuously splits the space into sub-regions and merges them based on the minimum distance from a plane. In the end, an octree structure is produced, which can deal with non-uniform sampled data but cannot handle coplanar patches that share borders. In addition, the octree operations are computationally expensive and do not scale very well with increased data. 

The authors in \cite{Bau2008} use an interesting variation where the Radon transform is computed for a grid of points and a density function is assigned to each sample. From that, a normal is calculated for each point and the latter  are clustered into distinct planes.

\cite{Got2004} extract planes and quadric surfaces from range image data with the help of a robust genetic algorithm-based estimator, looking at  the local orientation of 3D points and the iterative use of a simplistic planarity test. Finally,~\cite{Ken2008} use a two-step iterative method based on a collection of local geometric patterns to classify 3D cloud points into polyhedral, planar regions. Their method is robust to noise and small quantization errors and may be used as a pre-cursor for subsequent plane extraction.

\section{Probabilistic model\label{sec:Prob_model}}
In this section we present a probabilistic treatment of patch extraction, cast as a recursive classification / density estimation problem. We begin with an initial estimate of a number of patches $\Psi_i(\bUpsilon_i,\Pi_i)$ defined by the planes $\Pi_i(\overrightarrow{n}_i,S_i)$ and their respective, constituent 3D points $\bUpsilon_i$=$\{\Upsilon_1,...,\Upsilon_n\}$. The latter specify the closed, convex boundaries $V_i$=$\{\Upsilon_1,...,\Upsilon_m\}$ with $m\leq n$. These initial patch estimates (determined by a seeding step) form the fixed, finite set of classes $\{ \Psi_1,...,\Psi_c \}$.

We may then express the posterior probability of obtaining the patch $\Psi_i$ given the points $\bUpsilon_t$ as:
\begin{equation}
P(\Psi_{i}|\bUpsilon_t,h)=\frac{P(\bUpsilon_t|\Psi_{i},h)P(\Psi_{i}|h)}{\sum_{j=1}^c{P(\bUpsilon_t|\Psi_{j},h)P(\Psi_{j}|h)}}.\label{eq:General_Bayes}
\end{equation} 
The likelihood term $P(\bUpsilon_t|\Psi_{i},h)$ encapsulates our observations on the membership of $\bUpsilon_t$ to a specific patch $\Psi_i$, given the geometric properties of the latter and any additional background evidence $h$, such as camera configuration, noise model and so on. The likelihood consists of two distance measures: first is the distance $d_{V_i}$ of $\bUpsilon_t$ from $V_i$ and second the distance $d_{\Pi_i}$ of $\bUpsilon_t$ from the normalised plane $\Pi_i$:
\begin{equation}
d_{\Pi_i} = (A_iX_t+B_iY_t+C_iZ_t+D_i)^2. \label{eq:dP}
\end{equation}
The calculation of $d_{V_i}$ is more complicated and involves partitioning
an $m$-vertex polygon into $m$-$2$ coplanar triangles and calculating
the minimum squared distance between $\bUpsilon_t$ and each of the
triangles. These two distance measures will ensure that $\bUpsilon_t$ is both close to the plane but also not too far from the patch boundary. The latter is used to avoid prematurely merging coplanar patches that may be far away and possible belong to different objects.

In terms of probabilities, in the case of $d_{\Pi_i}$ we are only considering the perpendicular distance from the plane, so if the possible locations of $\bUpsilon_t$ relative to $\Pi_i$ can be approximated with a Gaussian distribution along the plane's normal, then as a result $d_{\Pi_i}\sim\Gamma(\gamma_1,\upsilon_1)$, which is a $\Gamma$ distribution of shape and scale parameters $\gamma_1,\upsilon_1$ respectively. For $d_{V_i}$, which is the sq. Euclidean distance, $\bUpsilon_t$ may be considered to have an approximate multivariate Gaussian distribution around a specific point on the patch, resulting in $d_{V_i}\sim\Gamma(\gamma_2,\upsilon_2)$.
 
As we mentioned earlier, we require $\bUpsilon_t$ to be both within a minimum distance from $\Pi_i$ and from $V_i$, so instead we consider the joint distance $d_i$=$d_{\Pi_i}$+$wd_{V_i}$, where $w$ is a weighting factor used to shift emphasis between the two distance measures. So for example when $w$$>$1 then $d_{V_i}$ will be up-scaled so we will reject what would otherwise be smaller distances from the patch boundary in favour for larger distances from the plane. The opposite occurs when 0$<$$w$$<$1.  Given our previous assumptions, the sum $d_i$ will have a distribution which is the convolution of $\Gamma(\gamma_1,\upsilon_1)\otimes\Gamma(\gamma_2,w\upsilon_2)$ and may be approximated to a good degree by $\Gamma(\alpha,\beta)$ where: $\alpha$=$\mu^2/\sigma^2$, $\beta$=$\sigma^2/\mu$ and $\sigma^2$=$\gamma_1\upsilon_1^2$+$\gamma_2w^2\upsilon_2^2$ and $\mu$=$\gamma_1\upsilon_1$+$\gamma_2w\upsilon_2$ are the variance and mean of $d_i$~\cite{Ste2007}. Therefore,
\begin{equation}
P(\bUpsilon_t|\Psi_{i},h)=P(\bUpsilon_t|d_i)=\frac{1}{\beta\Gamma(\alpha)}\left(\frac{d_i}{\beta}\right)^{\alpha-1}\exp\left(\frac{-d_i}{\beta}\right).
\label{eq:likelihood}
\end{equation}

Next is the prior term $P(\Psi_{i}|h)$=$P(d_{c})P(I,I')$, which
will incorporate our background knowledge about the geometry and greyscale intensity
of the scene and patch respectively. This term  is not dependent on the patch growth and so it may be calculated in advance during initialisation.
More specifically, $P(d_{c})$ is
a prior based on the induced properties and configuration of the cameras
and tracking errors that manifest as Gaussian noise in the two 2D
images. This prior models the uncertainty associated with the 3D reconstruction
and thus penalises points that are far away from both cameras and/or
outside the field of view. As such, $P(d_{c})$ will incorporate
the triangulation information and our confidence about the 3D location
of $\bUpsilon_t$.

We may formulate this as a pdf that assigns low probability to large
distances from $\bUpsilon_t$=$T(\bx_t,\bx_t',K,K')$, where $T$ is the triangulation
operation~\cite{HarStu1997} which projects the points $\bx_t,\bx_t'$ from the two cameras with
projection matrices $K,K'$ onto rays that intersect in 3D space.
In addition, we may consider $\tilde{\bUpsilon}_t$=$T(\bx_t$+$N(0,\sigma),\bx_t'$+$N(0,\sigma'),K,K')$
as the estimated 3D point under the Gaussian noise assumption in the
points in the two images. We may therefore penalise for large distances
$d_{c}$=$||\bUpsilon_t$-$\tilde{\bUpsilon}_t||^{2}$, which will occur
when there is high uncertainty in the stereo estimation (i.e. associated
reconstruction error. Equally, we may penalise for large
deviations of $d_{c}$ from zero, using the probabilistic model:
\begin{equation}
P(d_{c})= \frac{k}{\lambda}\left(\frac{d_{c}}{\lambda}\right)^{k-1}\exp \left(-\left(\frac{d_{c}}{\lambda}\right)^{k}\right)   ,d_{c}\geqslant0 .\label{eq:noise_prior}
\end{equation}
 This is the Weibull distribution and is used here as an approximation
to the actual distribution of $d_{c}$. More specifically, if the
noise is i.i.d. Gaussian distributed in the two images, then its magnitude
will be $\Gamma$ distributed. However, due to the mutual perspective
projections, the probability $P(\bUpsilon_t|\bx_t,\bx_t')$ of obtaining the 3D point $\bUpsilon_t$ given the 2D points $\bx_t,\bx_t'$ in the two images (see Fig. \ref{fig:Real_results} (b)),  will not be strictly
Gaussian and so $P(d_{c})$ is not strictly $\Gamma$, but only in cases
of little or moderate uncertainty. Our experiments have shown that
in all cases the Weibull in (\ref{eq:noise_prior}) provides an equal
or better fit than the $\Gamma$, with both models being equivalent when
$P(\bUpsilon_t|\bx_t,\bx_t')$ is approximately Gaussian. Note that neither of
the two models provide an exact fit to the data under extreme uncertainty
conditions when the variance of $P(\bUpsilon_t|\bx_t,\bx_t')$ approaches infinity. The parameters $k,\lambda$ from (\ref{eq:noise_prior}) remain fixed throughout the patch extraction process and are estimated based on ground-truth training samples.

The reason why we use $P(d_{c})$ to penalise for uncertainty is to avoid
highly erroneous points affecting the fitment of the plane and thus
(in the process) reject many good points which would otherwise be
included in the patch. It is preferable to reject a very uncertain
point than to adapt the plane in order to fit it. In this way, the
overall quality of the patch is maintained.

Finally, $P(I,I')$ is a prior that represents an acceptance - rejection decision based on the assigned intensity to $\bUpsilon_t$ from its corresponding 3D point pair $\bx_t,\bx_t'$ in the two images. The intensity values are given by the initial segmentation (see Sec. \ref{subsec:seeding}) and are used as an approximate 3D patch pre-segmentation step. A point $\bx$ should have a high probability when close to the centre of its assigned segment, with the probability falling off smoothly near the segment's elliptical boundary. This smooth falloff is necessary because in the 2D image, a recovered elliptical segment is only a crude approximation to the shape of the segmented region.

Using therefore the segmentation method by~\cite{ForMoe2006} we have effectively converted from
intensity information to a geometric representation. In other words,
different grayscale patches in the image may be represented by ellipses
of constant intensity and their geometric properties (position, orientation
and scale) easily modelled by a bivariate Gaussian distribution. This
approximation of course is more valid for images of objects of elliptical
shape and of near-constant pixel intensity. Objects with complicated
texture will give rise to many small ellipses and thus numerous patch
seed points, although this may be suppressed by the segmentation algorithm.

For this prior, we will align two bivariate Gaussians with the given
ellipses. Assuming independence in the two images we get $P(I,I')$=$P(I)P(I')$, and given an ellipse defined by its centroid $M_i$=$(m_{\chi},m_{\eta})$ and
inertia matrix $\Xi_i$=$\left[\begin{array}{cc}
k_{1} & k_{2}\\
k_{2} & k_{3}\end{array}\right]$ we set:
\begin{equation}
P(I)=\frac{1}{2\pi\sqrt{(1-\rho^{2})k_{1}k_{3}}}\exp\left[-\frac{z_t}{2(1-\rho^{2})}\right],\label{eq:intensity_prior}
\end{equation}
where: $z_t=\frac{(\chi_t-m_{\chi})^{2}}{k_{1}}-2\rho\frac{(\chi_t-m_{\chi})(\eta_t-m_{\eta})}{\sqrt{k_{1}k_{3}}}+\frac{(\eta_t-m_{\eta})^{2}}{k_{3}}$, $\bx_t$=$(\chi_t,\eta_t) \sim K\bUpsilon_t$ is the back-projection of $\bUpsilon_t$ onto image $I$
and $\rho$=$\frac{k_{2}}{\sqrt{k_{1}k_{3}}}$ is the correlation coefficient.
Similarly for $P(I')$ with $M_i'$.

\subsection{Classification}
Given therefore the set  $\bUpsilon$  of all points in the scene, and an initial estimation of the parameters of the posterior (\ref{eq:General_Bayes}) we need to determine the formers' class memberships. We may thus define a set of discriminant functions $g_i(.)$, $i$=$1,...,c$. The classifier will then assign a point $\Upsilon \in \bUpsilon$ to class $\Psi_i$ iff:
\begin{equation}
g_i(\Upsilon)>g_j(\Upsilon)>\tau \  \textnormal{for all} \  j\neq i , \label{eq:general_discriminant}
\end{equation}
where $\tau$ is some threshold to avoid classifying excessively noisy points. We thus compute $c$ discriminant functions for each point and assign that point to the largest discriminant if the latter is above some threshold. Assuming that all misclassification errors are equally costly, (\ref{eq:general_discriminant}) becomes:
\begin{equation}
P(\Psi_i|\Upsilon) > P(\Psi_j|\Upsilon) > \tau \  \textnormal{for all} \  j\neq i . \label{eq:Bayes_discriminant}
\end{equation}
We may proceed further by taking the log of (\ref{eq:Bayes_discriminant}), which does not change the results of the classification. As such, from (\ref{eq:Bayes_discriminant}) we have:
\begin{equation}
\log P(\Psi_i|\Upsilon,h) > \log P(\Psi_j|\Upsilon,h) > \tau_L \  \textnormal{for all} \  j\neq i , \label{eq:log_discriminant}
\end{equation}
and 
\begin{equation}
\log P(\Psi_{i}|\Upsilon,h)  \propto (\alpha -1)\log(d) -\zeta\left( \frac{z}{2\omega}+\frac{z'}{2\omega'} \right)-\frac{d}{\beta}+C_1 ,
\label{eq:log_probability}
\end{equation}
where $C_1$=-$\alpha \log(\beta)$ - $\log(\Gamma(\alpha))$ - $\frac{1}{2} \log(\omega \omega' k_1 k_3 k_1'k_3')$
is constant for a given $\Psi_i$. Also
\begin{equation}
\tau_L=\log(\tau) + \underbrace{( \frac{d_c}{\lambda})^k - (k-1)\log(d_c)}_{\text{$F(\Upsilon)$}}  +C_2, \label{thresh_tau}
\end{equation}
where 
$C_2$=$k\log(\lambda)$ -$ \log(k)$
is constant for all $\Upsilon$. $F(\Upsilon)$ is not important for the discriminant comparison but is required for the thresholding and it may be precomputed for all $\Upsilon$ in advance. Here we set $\omega$=$1$-$\rho^2$ and $\omega'$=$1$-$\rho'^2$. Finally, we note the weight $\zeta$ which is used to adjust the effects of the prior distributions $P(I),P(I')$ depending on the complexity of data. So for example, when our data is replete with multi-textured objects, and strong perspective effects we might wish to reduce the input coming from the 2D segmentation algorithm and give more emphasis on the 3D geometry instead.

\subsection{Density estimation}

\begin{figure}
\centering
		\begin{tabular}{cc}
				\includegraphics[width=0.15\textwidth]{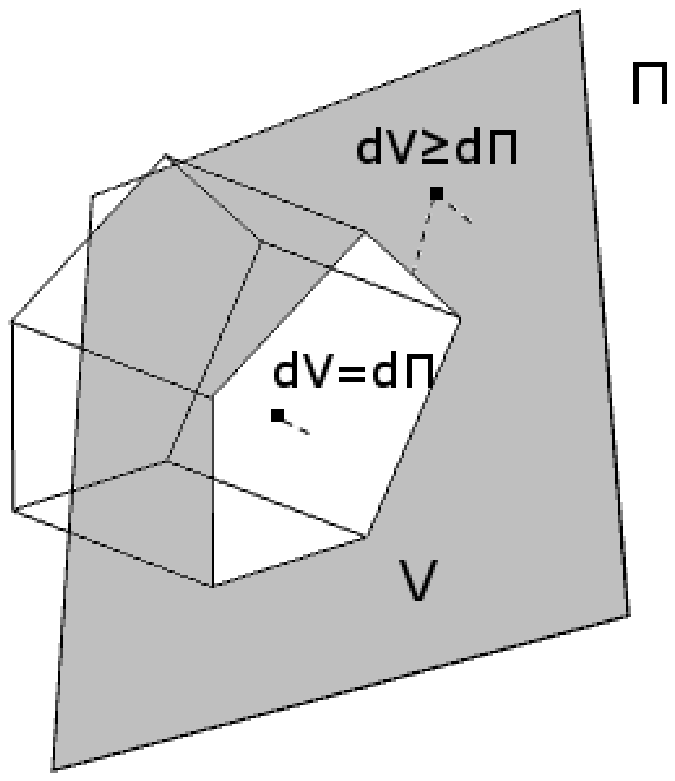} &
      	\includegraphics[width=0.19\textwidth]{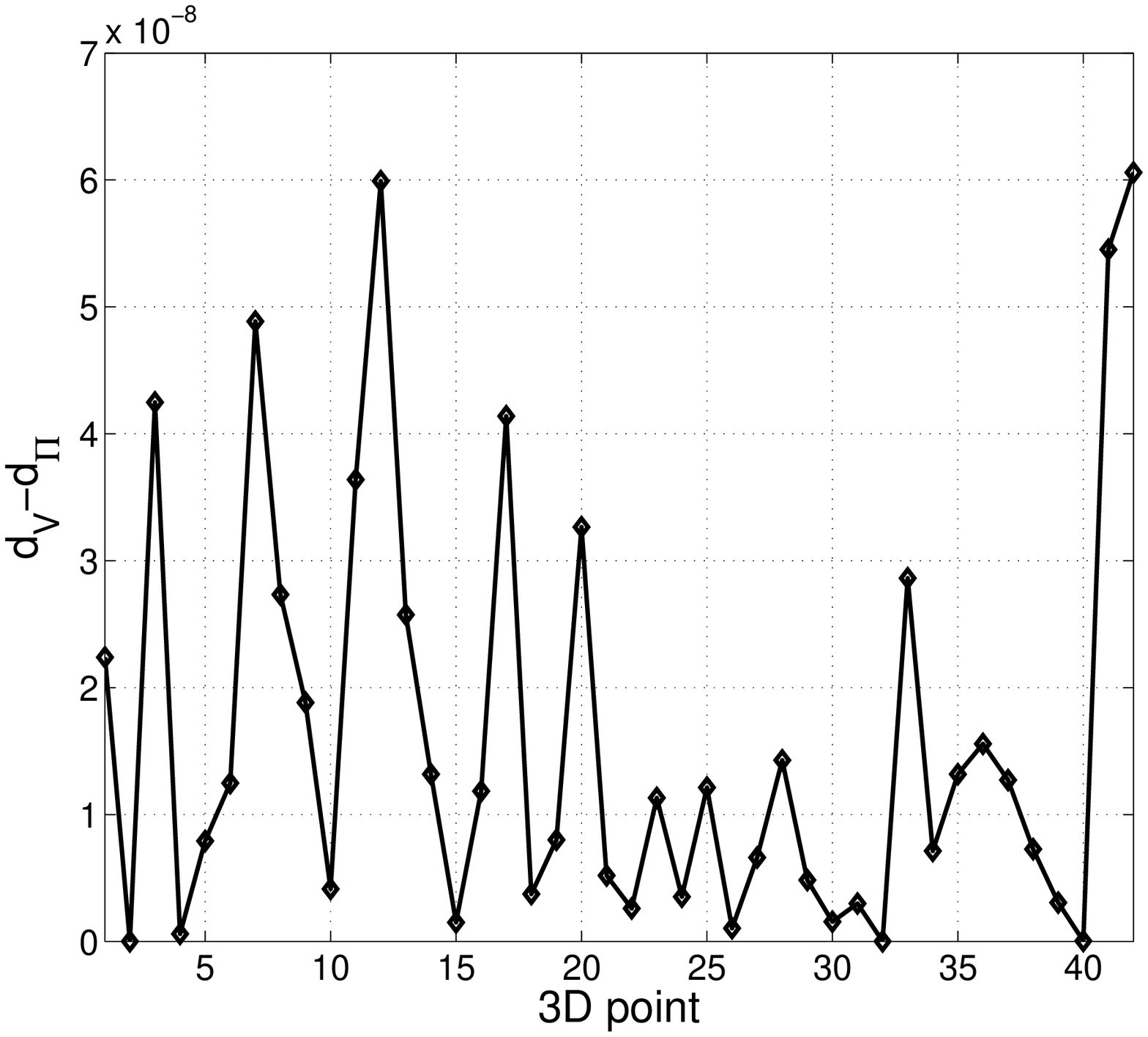} \\
      	(a) & (b)
		\end{tabular}

	\caption{The two distance measures $d_V$ and $d_\Pi$ (a) and their typical difference for points in a patch $V$ (b).}
	\label{fig:half_space}
\end{figure}

Once we have assigned a dataset $\Omega_i$=$\{ \Upsilon_1,...,\Upsilon_n \}$ to a class $\Psi_i$ using (\ref{eq:log_discriminant}), we require a new estimate of the density varying parameters $\theta_i$=$(\alpha_i,\beta_i)$. This resembles an iterative training of the classifier, where we can update our knowledge about $\theta_i$ given $\Omega_i$. Assuming that we have no explicit prior knowledge about $\theta_i$ then we can define the likelihood w.r.t. the set of samples as:
\begin{equation}
P(\Omega_i|\theta_i)=\prod_{t=1}^n{P(\bUpsilon_t|\Psi_i,\theta_i)}=\prod_{t=1}^n{P(\bUpsilon_t|d_i,\theta_i)}, \label{eq:likelihood_estimate}
\end{equation}
where $\bUpsilon_t$ are i.i.d.. The MLE $\hat{\theta}_i$ may be obtained by taking the log-likelihood of (\ref{eq:likelihood_estimate}) and setting the derivative w.r.t. $\theta_i$ to zero:
\begin{equation}
\sum_{t=1}^n{\nabla \theta_i \log P(\bUpsilon_t|d_i,\theta_i)}=0. \label{eq:MLE}
\end{equation} 
For the $\Gamma$ distribution, there are known ML approximations of the parameters:
\begin{equation}
\hat{a}_i \approx \frac{3-\phi_i + \sqrt{(\phi_i-3)^2+24\phi_i}}{12\phi_i}, \: \hat{\beta}_i \approx \frac{1}{\alpha_i N}\sum_{t=1}^n{d_{i_t}} \label{eq:MLE_ab},
\end{equation}
with $\phi_i$=$ \log(\frac{1}{n}\sum_{t=1}^n{(d_{i_t}})$-$\frac{1}{n}\sum_{t=1}^n{\log (d_{i_t})}$. As a result of (\ref{eq:MLE_ab}), there is the potential for an efficient iterative implementation, since $d_{\Pi_i}$ in $d_i$ may be calculated simultaneously for all points in $\bUpsilon_t$ using a matrix multiplication, and $d_{V_i}$ usually only involves a small subset of points. We can also increase computation speed by exploiting the fact that when the points are inside the half-space defined by the patch convex boundary (Fig. \ref{fig:half_space}), then $d_{\Pi_i} \approx d_{V_i}$ and so $d_i\approx 2wd_{\Pi_i}$ without significant loss of accuracy. When a point is outside the half-space then usually $d_{\Pi_i} \neq d_{V_i}$ so we need to calculate both distances for equation (\ref{eq:likelihood}).

\begin{figure}
\centering
				\includegraphics[width=0.23\textwidth]{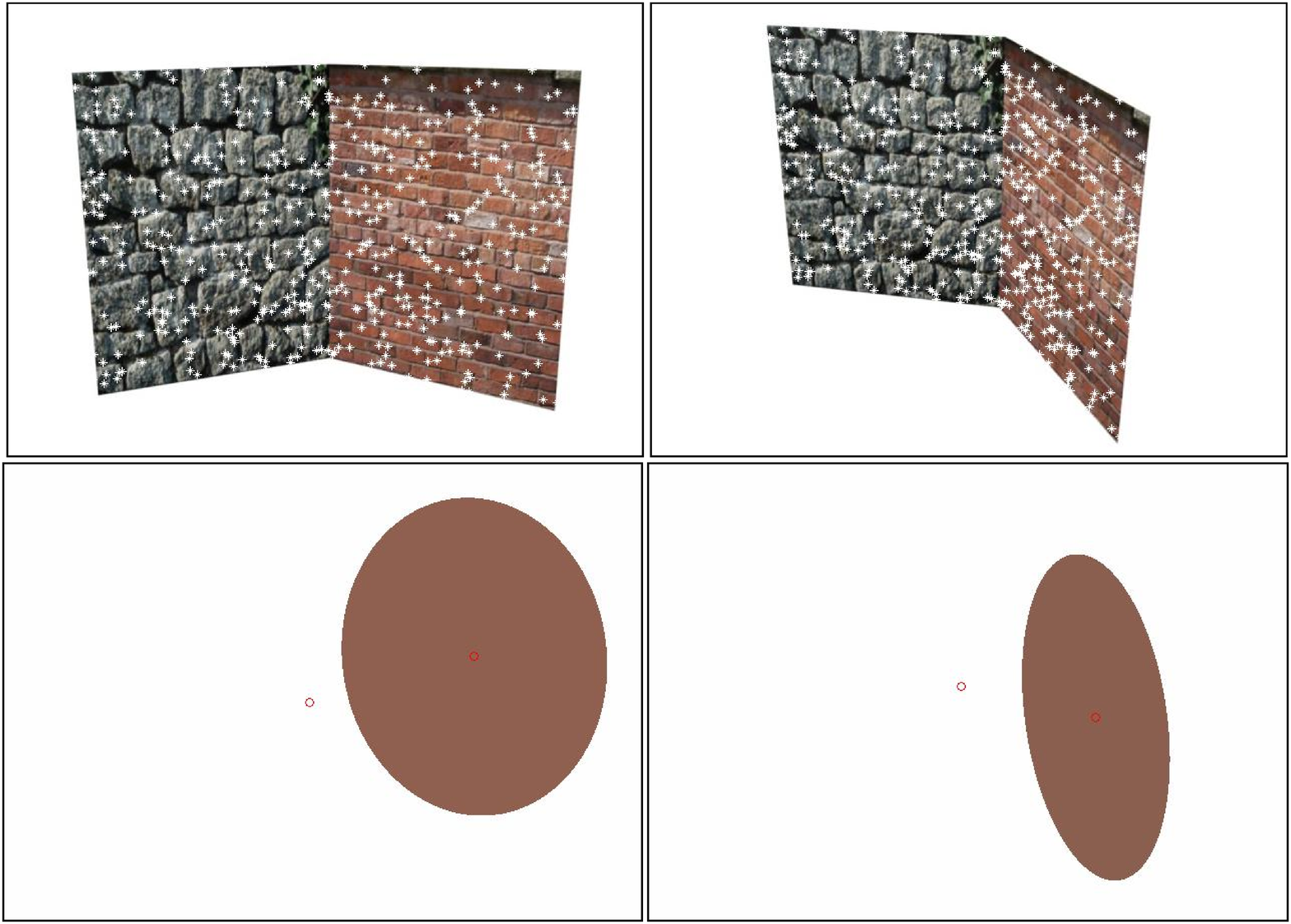} 
				\includegraphics[width=0.23\textwidth]{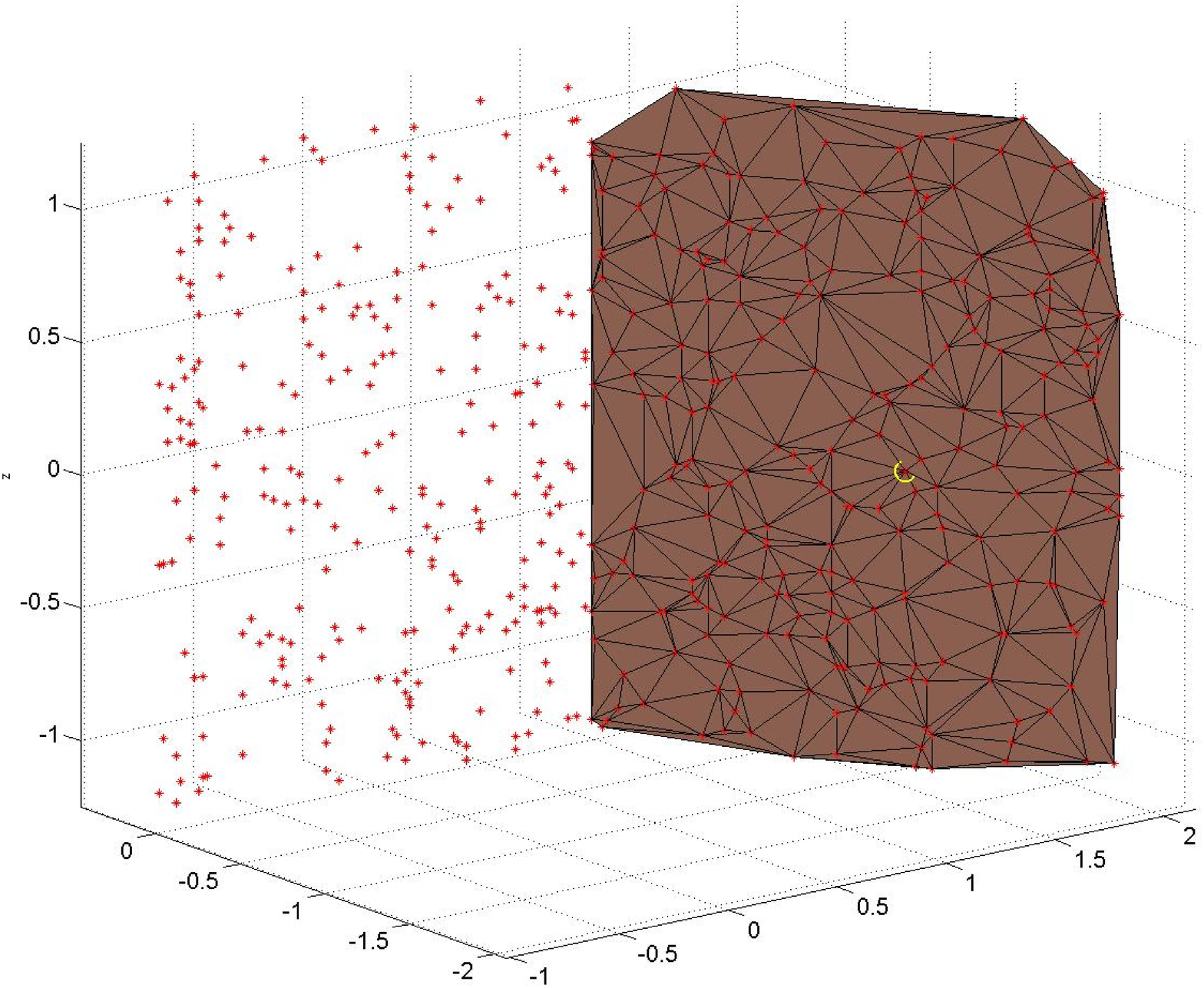} 
	\caption{Simple example. (Upper left) Two views with superimposed interest points. (Lower left) 2D segmentation. (Right) Extracted planar patch in 3D. }
	\label{fig:Example}
\end{figure}

\section{Patch extraction\label{sec:Algorithm}}
In this section we summarise the proposed probabilistic inference model in terms of a compact algorithm suited for efficient computational implementation.

\subsection{Seeding \label{subsec:seeding}}
This is the first step of our algorithm, and will determine the initial plane equation from which the patch will grow and evolve over time. Although the plane equation will adapt to any new points added, the initial seed point is significant since a poor choice will most likely cause the growth to end prematurely and/or extract a meaningless planar patch.

Our assumption is that structurally and perceptually meaningful patches are those that segment the 3D scene into distinct objects and coincide with planar and linear boundaries. Since object topology and boundary information in a 3D scene is largely preserved in a 2D image of the scene, we have employed a 2D region segmentation algorithm~\cite{ForMoe2006} in order to automatically determine the initial, seed points of our patches. We chose this particular segmentation solution since it provides an efficient approximation of segments with ellipses that can translate very easily into probability distributions. In addition, it can achieve a consistent and corresponding segmentation across the two views.
 The output of this algorithm is therefore a corresponding partition of the images $I,I'$ into elliptical areas of constant colour intensity (see Fig. \ref{fig:Example}). The centres of these ellipses are pair-wise triangulated in 3D~\cite{HarStu1997} and used as the seed points for an equal number of patches. We denote a seed point as $S_i$=$T(M_i,M_i',K,K')$. It is however the case that  since an ellipse is not perspective invariant, this method may fail to produce correctly aligned 3D ellipsoids and centroids that triangulate exactly onto the 3D mesh, when we have strong perspective effects in the two scene views. In this case, we may downweigh the effects of the segmentation as described in (\ref{eq:log_probability}).

Once the seed points have been defined, we need to select adjacent points on which to fit the initial 3D planes. To do this, we select the 3D points inside the spheres with centres $S_i$ and radius some pre-defined threshold $R_\tau$ relative to the volume or density of the 3D cloud. We then fit a plane $\Pi_i$ to each of these spherical clusters using the slope-intercept form of the plane, for example $Z$=$aX$+$bY$+$d$, so that we only have 3 unknown coefficients but also the subsequent equations are linear. This approach assumes that our measurements in one component are functionally dependent on the other two, so in this case the $Z$-component is functionally dependent on the $X$- and $Y$-components of the sample points  $(X_t,Y_t,f(X_t,Y_t))$. In addition, any noise is distributed in the $Z$-direction.

Given the above assumptions, we seek the 3 coefficients so that the fitted plane minimises the sum of squared errors in the $Z-$direction. We may define the error function as
\begin{equation}
E_{a,b,d}=\sum[aX_{t}+bY_{t}+d-Z_{t}]^{2}, \label{plane_fit}
\end{equation} 
the minimum of which occurs when the gradient satisfies $\nabla E_{a,b,d}$=$(0,0,0)$. This leads to a system of three linear equations in $a,b,d$:
\begin{equation}
\nabla E_{a,b,d}=2\sum[aX_{t}+bY_{t}+d-Z_{t}](X_{t},Y_{t},1),  \label{plane_fit_deriv}
\end{equation} which becomes:
\begin{equation}
\left[\begin{array}{lll}
\sum X_{t}^{2} & \sum X_{t}Y_{t} & \sum X_{t}\\
\sum X_{t}Y_{t} & \sum Y_{t}^{2} & \sum Y_{t}\\
\sum X_{t} & \sum Y_{t} & \sum1\end{array}\right]\left[\begin{array}{c}
a\\
b\\
d\end{array}\right]=\left[
\begin{array}{l}
\sum X_{t}Z_{t}\\
\sum Y_{t}Z_{t}\\
\sum Z_{t}
\end{array}\right].
\label{eq:linear_system_plane}
\end{equation}
(\ref{eq:linear_system_plane}) is easily and quickly solved by inverting the $3\times3$ matrix and all that is required is calculation or update of the running sums. Similar equations to (\ref{eq:linear_system_plane}) result for the other two plane types $X$=$bY$+$cZ$+$d$ and $Y$=$aX$+$cZ$+$d$. 

Note that the slope-intercept fit only works when the slope of the plane is not too large, in which case we need to switch to one of the other two plane types. To determine which of the three types is best suited for our data sample, we simply look at the minimum range at each dimension in the spherical clusters of sampled points. The points that fit the plane equations will form the initial patches $\Psi_i$, while the outliers will be rejected and need not be considered by the same patch again. This slope test is carried out only once during seeding, since a patch is unlikely to change slope considerably during growth. This way, the slope-intercept method is much faster than the homogeneous fitting method with 4 coefficients. 

After the seeding stage, we have the initial planes $\Pi_i$ and can calculate the convex boundaries $V_i$ of the patches $\Psi_i$. In addition, we obtain the first estimates of $\theta_i$ using (\ref{eq:MLE_ab}).

\subsection{Growing}
Given an appropriate patch seed, we now consider a random point $\Upsilon_{n+1}$from the available 3D cloud \emph{not} belonging to any of the other patches \emph{nor} having being rejected already by the current patch. We denote such points as \emph{inliers} or in other words ``available for consideration''. Any points rejected by one patch may be re-considered later when the patch has grown, or by other patches. On the other hand, points already belonging to a patch are not be re-considered by any patch.

Therefore, we calculate a discriminant function (\ref{eq:log_probability}) for each patch $\Psi_i$ and assign $\Upsilon_{n+1}$ to the patch with the highest probability provided it is above some minimum threshold as in (\ref{eq:log_discriminant}). If the point is accepted, then it is removed from the inlier queue, otherwise it is returned on the top of the queue for later re-consideration.

The accepted point $\Upsilon_{n+1}$ is added to the current patch $\Psi_i$ dataset $\Omega_i$, which is used to update the current parameter estimate $\theta_i$. This is achieved by first fitting the plane $\Pi_i$, by updating the sums and solving (\ref{eq:linear_system_plane}). The planar convex hull $V_i$ may then be updated using any available, efficient algorithm (see~\cite{SchEbe2003} $\S$ 13.7.1). 

We then calculate the distances $d_i$=$d_{\Pi_i}$+$wd_{V_i}$ for all $\Upsilon \in \Omega_i$. $d_{\Pi_i}$ are calculated simultaneously from (\ref{eq:dP}) for all points using a matrix multiplication. Note that to convert from say, the X-slope-intercept form, to the normalised implicit form in (\ref{eq:dP}) we set: $A$=$1/ \sqrt{1+b^2+c^2}$, $B$=-$Ab$, $C$=-$Ac$ and $D$=-$Ad$. Equivalently for the other two plane types. $d_{V_i}$ is calculated  as described in Sec. \ref{sec:Prob_model}.

Following that we can use (\ref{eq:MLE_ab}) to obtain an updated estimate for $P(\Omega_i | \theta_i)$ for each patch. The growth process is repeated until termination, that is, when an already rejected point is revisited and there has been no change in $\Omega_i$ for all $i$. 
This whole process can be made very efficient, especially if the points are sorted in advance in order of descending joint distance from the two cameras, so that nearby points are checked first with a smaller chance of rejection. Additionally, under the assumption that the patches do not change considerably after a single-point classification / estimation step, we may increase the number of points considered concurrently at each step, in order to increase execution speed, without overly diverging from the original solution.

\subsection{Refinement}
The final step of the algorithm is the refinement stage where approximately similar, adjacent patches are merged together. In addition any degenerate patches with very few points or small areas are discarded. The merging process uses simple geometric and colour intensity information to determine if a pair of patches (at each time) should be combined.

The geometric consistency test uses two criteria, the first being the dot product of the two patches' normals $E$=$(\overrightarrow{n}_{i}\cdot\overrightarrow{n}_{j})$. This determines whether or not the two patches have approximately the same planar equation, but as this is not enough by itself,  we use the minimum squared distance between the two patches' convex hulls $V_i$ and $V_{j}$. For this, we make a simple distance check between two 3D polygons, which involves checking the minimum distance between triangle pairs. Merging the patches in this fashion results in a piece-wise planar patch that can cater for moderate amounts of noise or imperfections in the data, assuming that the distance thresholds have been set appropriately for the specific dataset.

The algorithm is shown in pseudocode in Alg. \ref{alg:algorithm} and a simple example is illustrated in Fig. \ref{fig:Example}. Note that we intentionally extracted only one of the two planar patches in this example, in order to demonstrate that the chosen patch will not grow over to the neighbouring region, partially biased by the segmentation prior and the joint distance likelihood. The result is a clean, distinct patch boundary, without any spikes or misclassified points, which distinguishes between the two, perceptually different regions. This is more difficult to achieve using only sparse geometrical information.
\begin{algorithm}
 Pick points $\bUpsilon_i$ inside spheres $(R_\tau,S_i)$
 
 Calculate $\Psi_i( \bUpsilon_i, \Pi_i ) , V_i , \theta_i $ \;
\Repeat {$\Upsilon_{n+1}$ is revisited without changes to $\Psi_i, \  \forall i$}
{
$\Upsilon_{n+1}$= bottom of inlier queue

\ForEach{patch $\Psi_i$} 
{
 $W$=$\max[ \log(P(\Psi_i|\Upsilon_{n+1},h))]$
}

\eIf{$W \geq \tau_L$} 
{
         Update $\Psi_i( \bUpsilon_i, \Pi_i) , V_i $ \;    
         Estimate $\theta_i$ \;
         Remove $\Upsilon_{n+1}$ from queue \;
}{
				 Re-insert $\Upsilon_{n+1}$ at top of queue

} 
}

 Refine $\Psi_i  \  \forall i$
 \caption{Patch extraction pseudocode.} \label{alg:algorithm}
\end{algorithm}

\section{Experiments\label{sec:experiments}}

\begin{figure*}
 	\centering
		\begin{tabular}{ccccc}
				\includegraphics[width=0.17\textwidth]{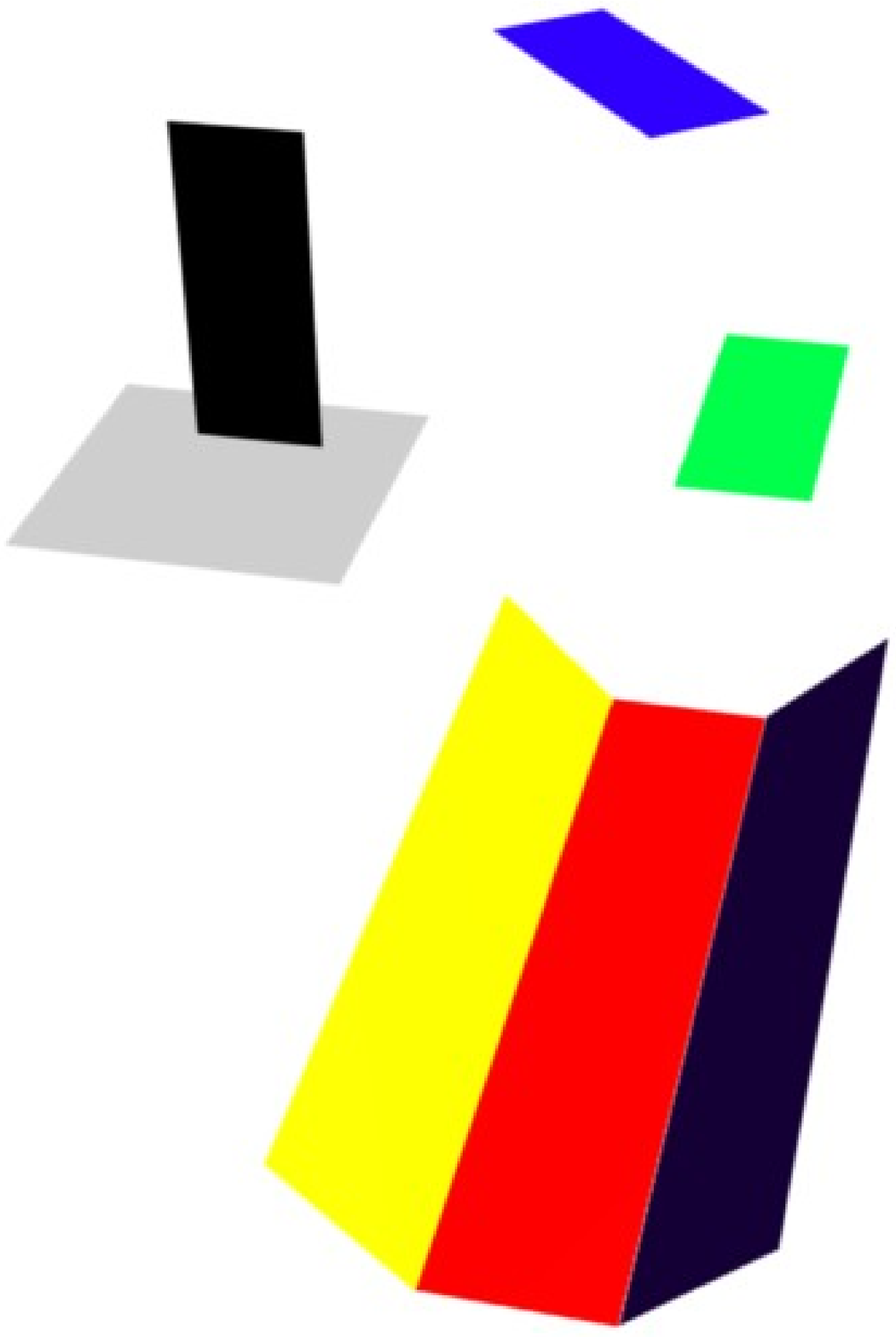} &
				\includegraphics[width=0.17\textwidth]{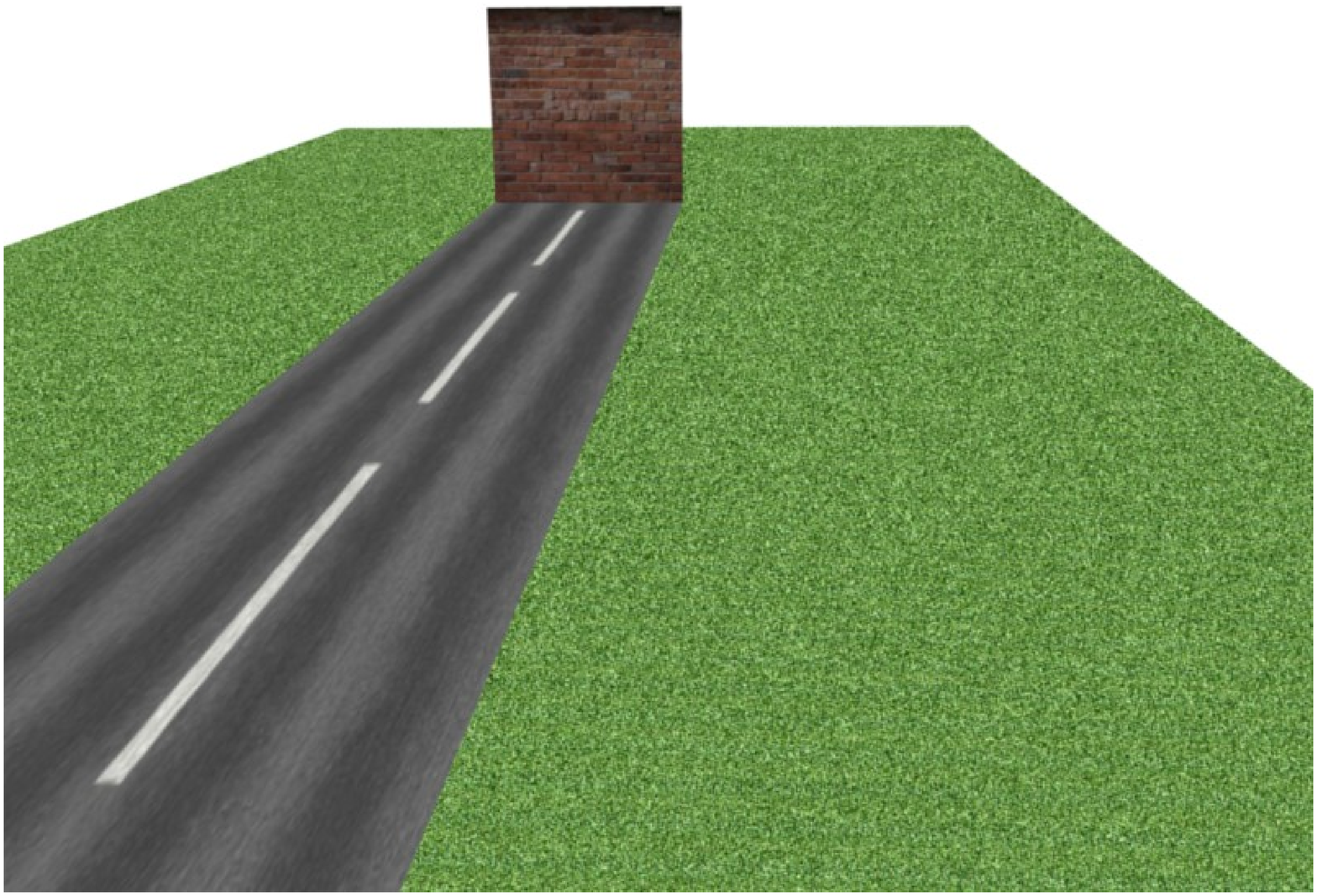} &
				\includegraphics[width=0.17\textwidth]{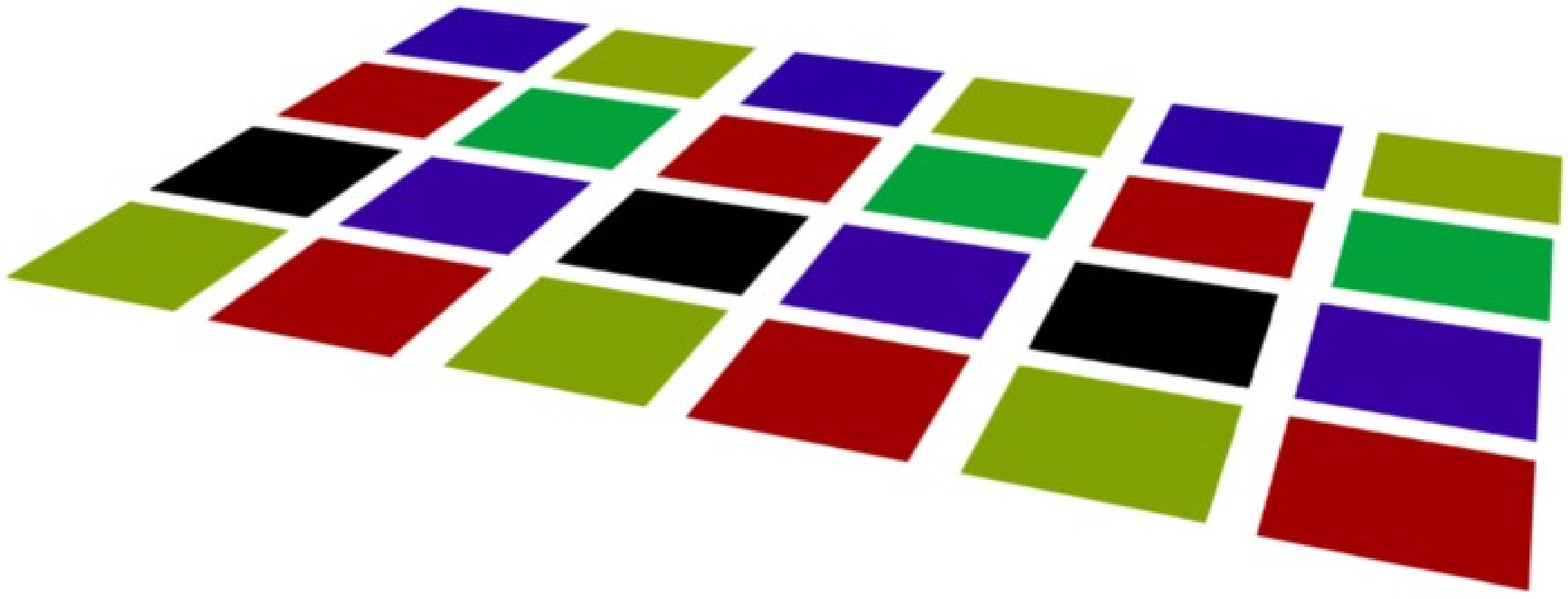} &    	
      	\includegraphics[width=0.17\textwidth]{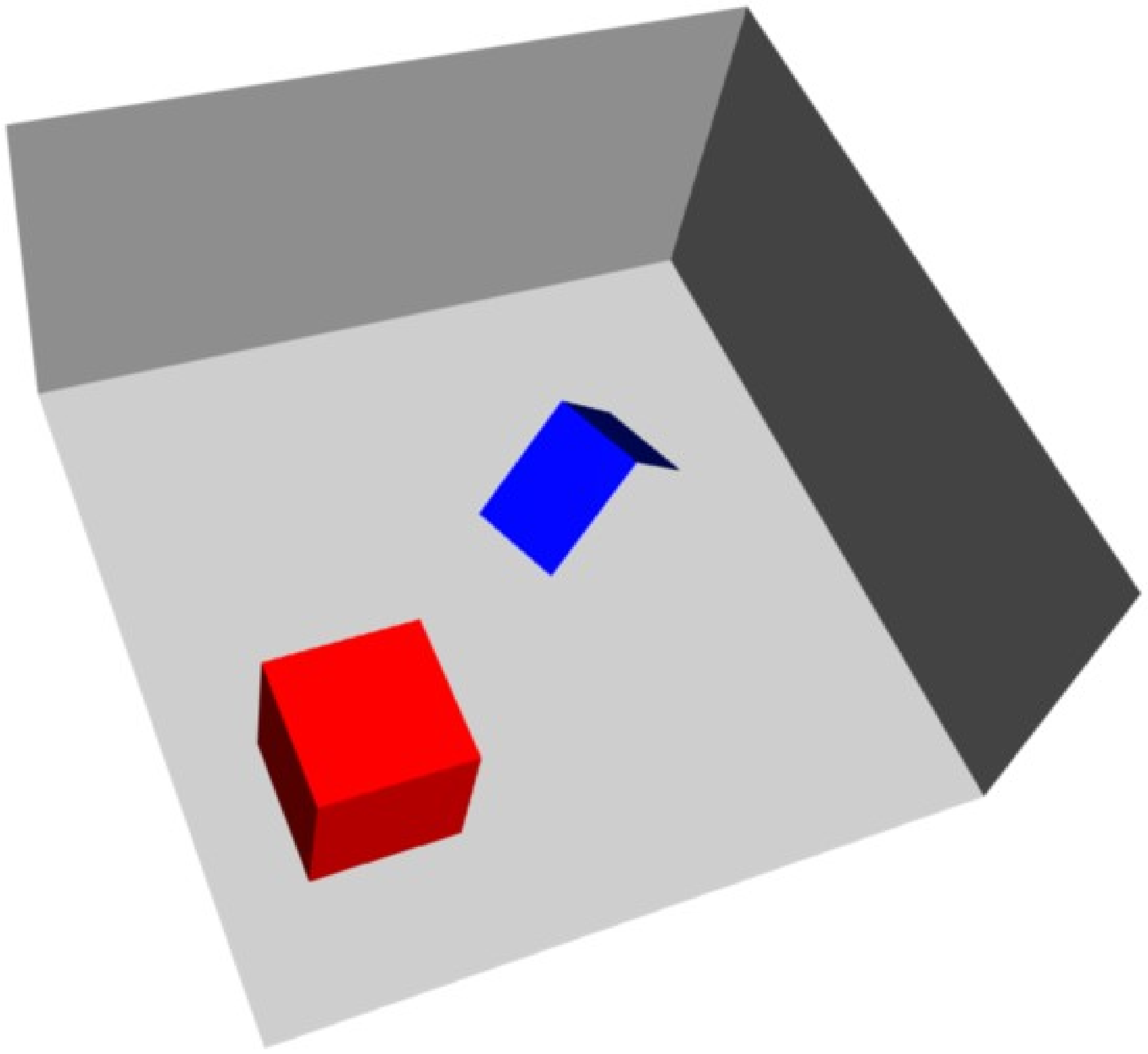} &
      	\includegraphics[width=0.17\textwidth]{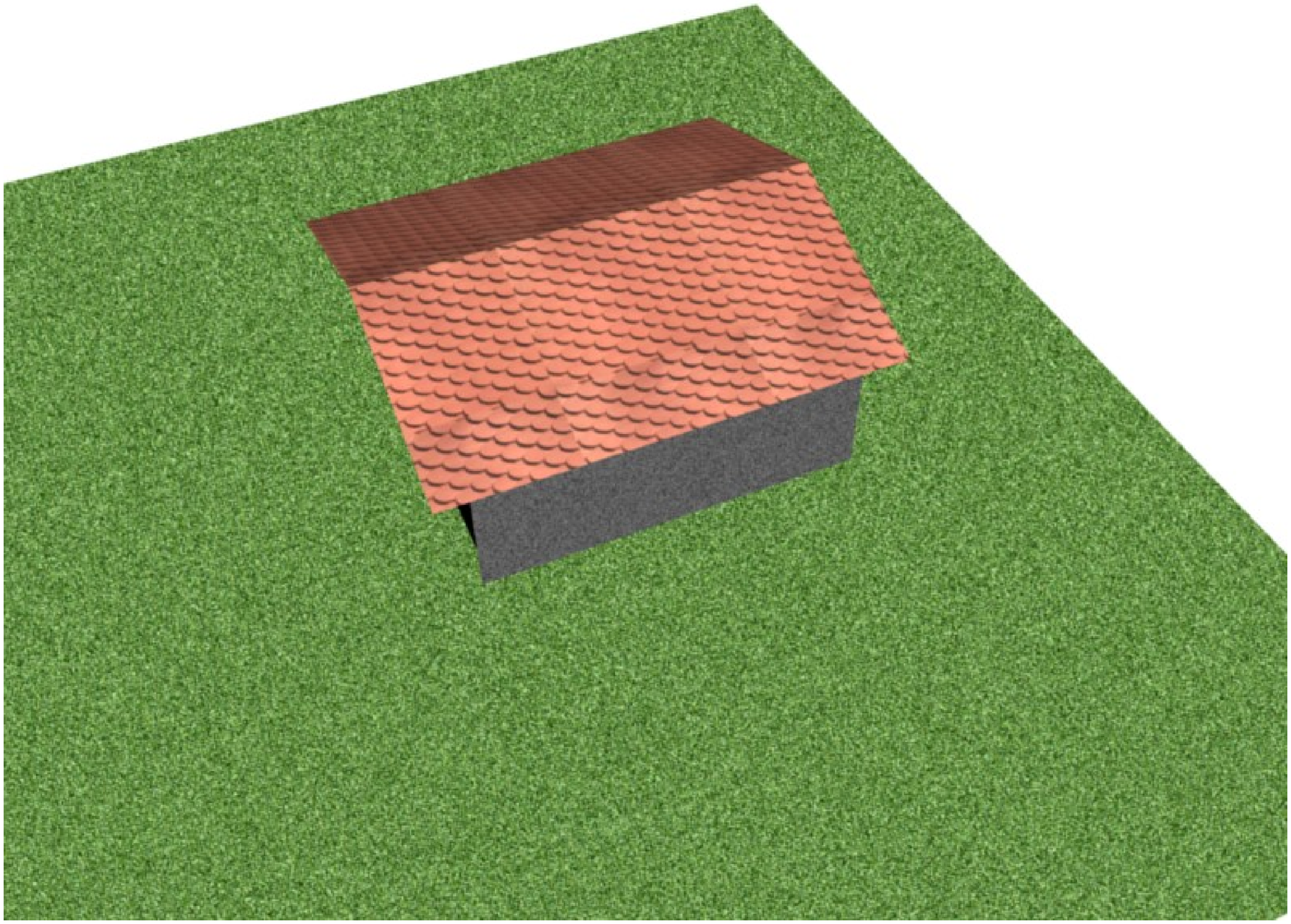} 
      	\\
      	\includegraphics[width=0.17\textwidth]{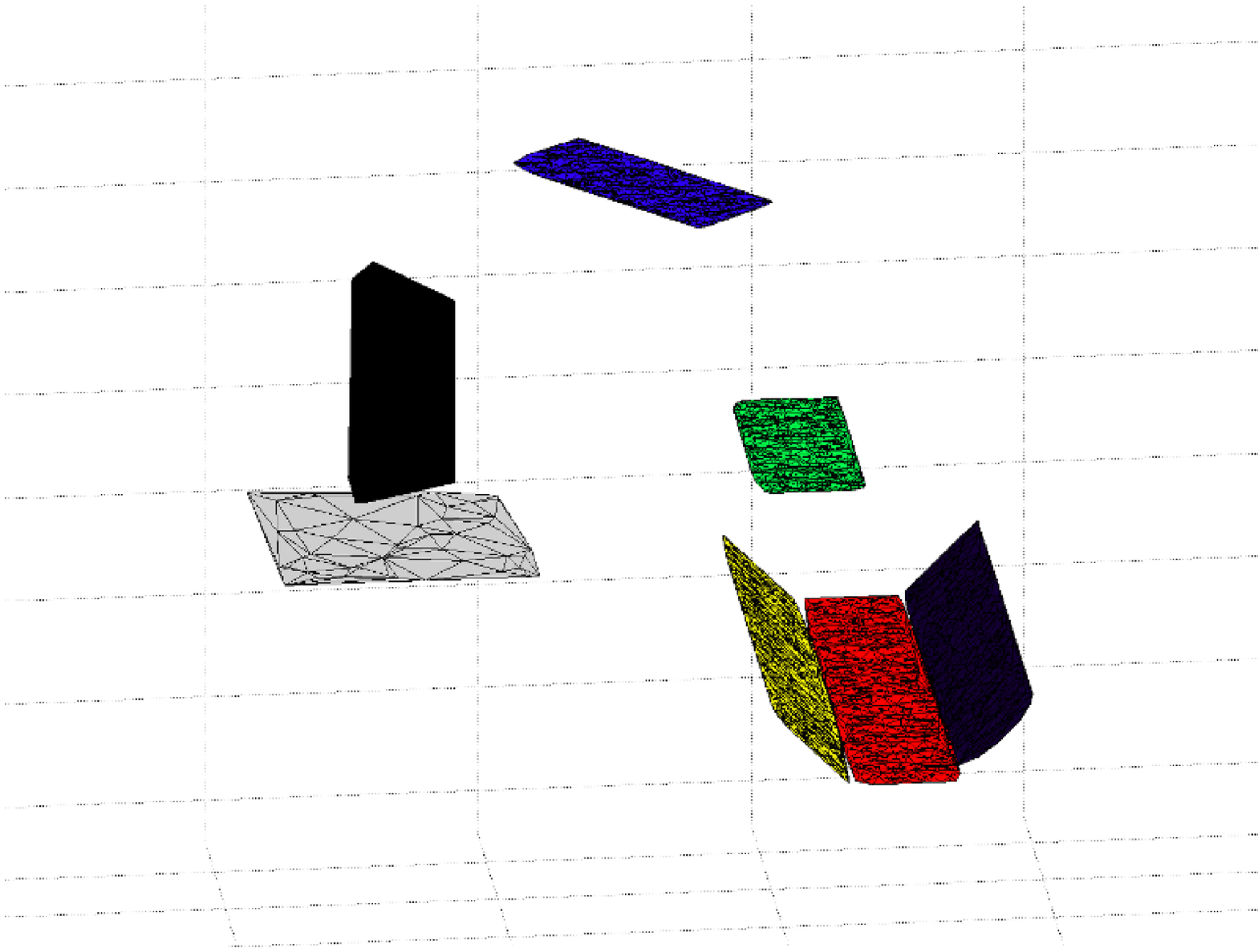} &
				\includegraphics[width=0.17\textwidth]{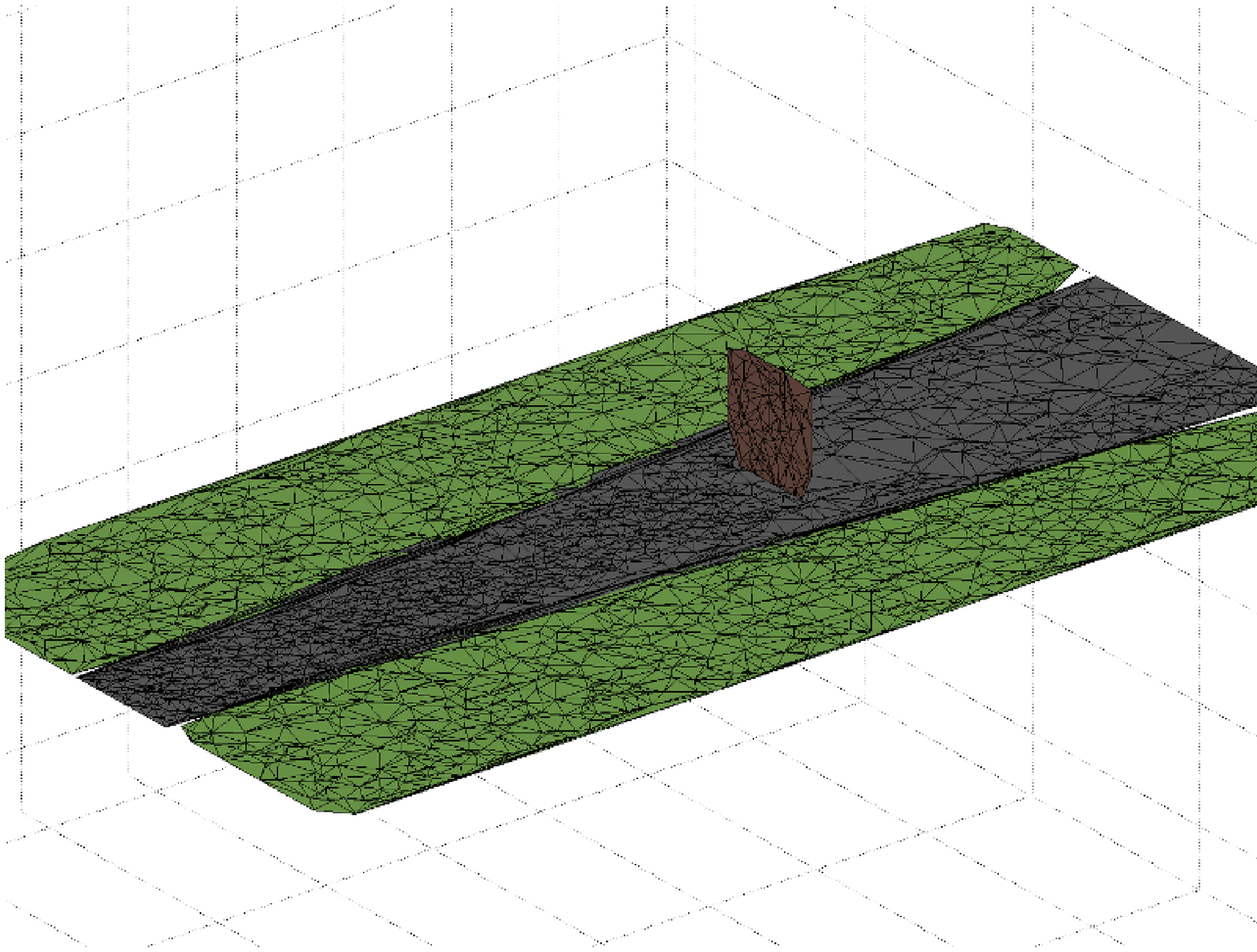} &
				\includegraphics[width=0.17\textwidth]{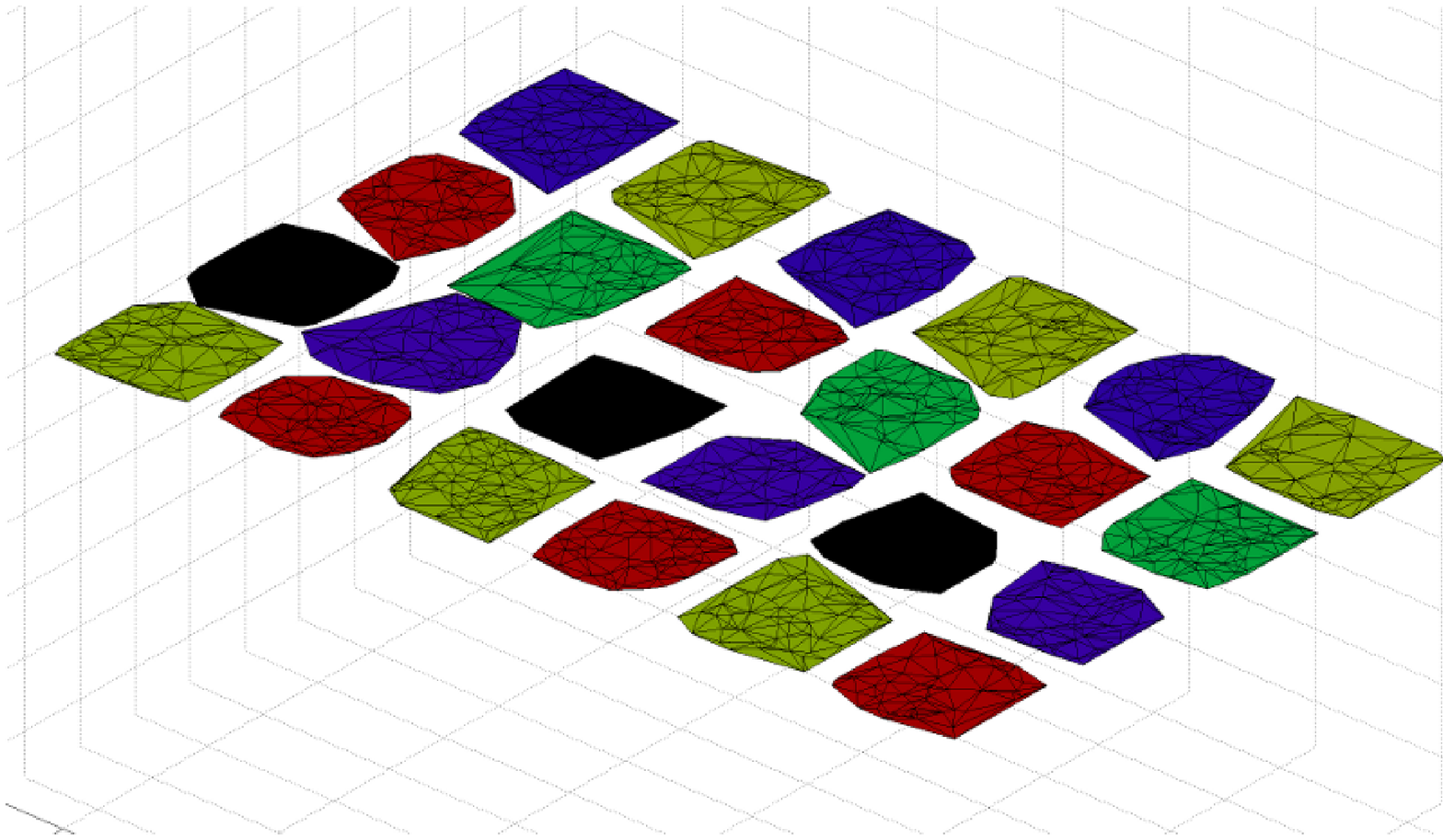} &    	
      	\includegraphics[width=0.17\textwidth]{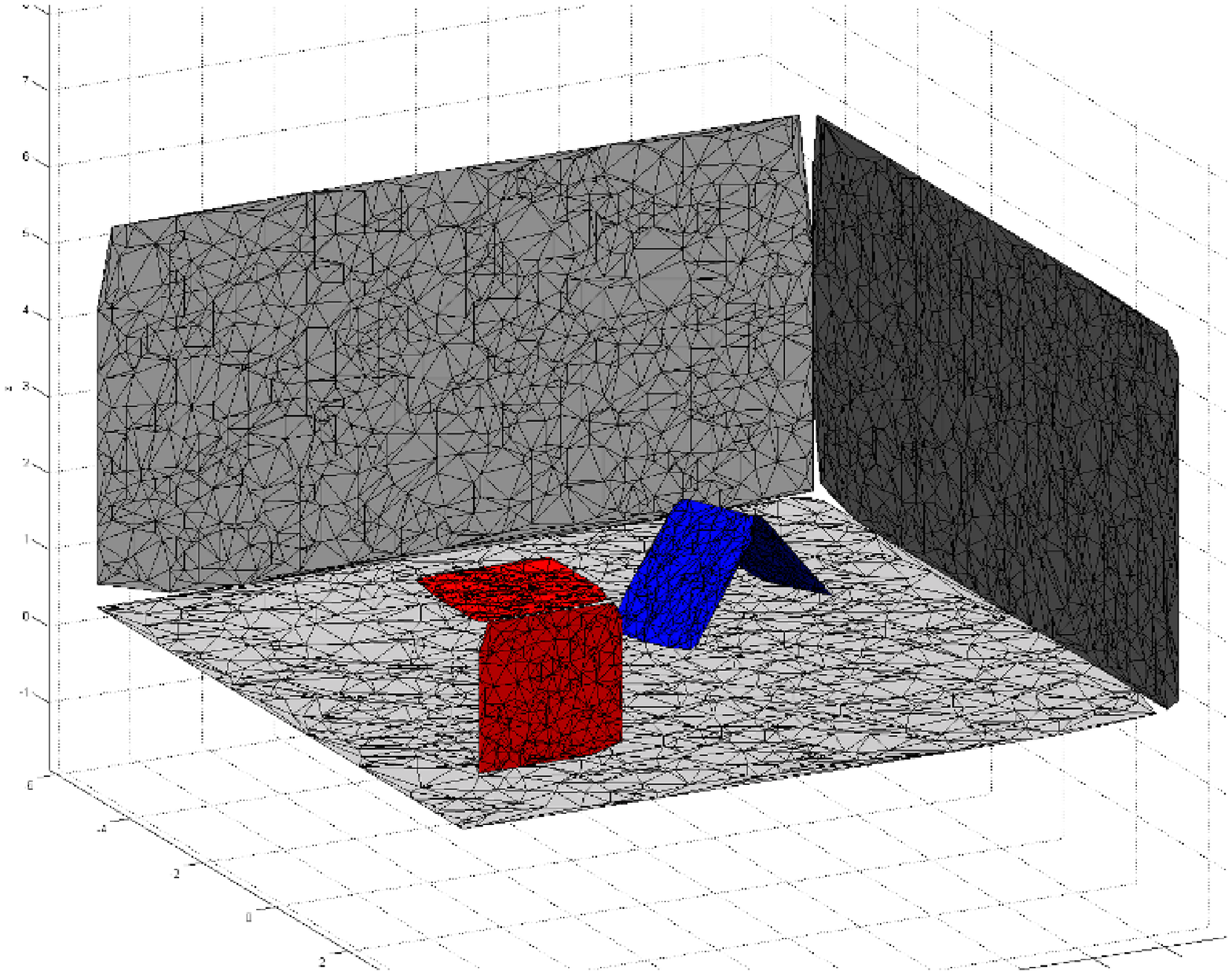} &
      	\includegraphics[width=0.17\textwidth]{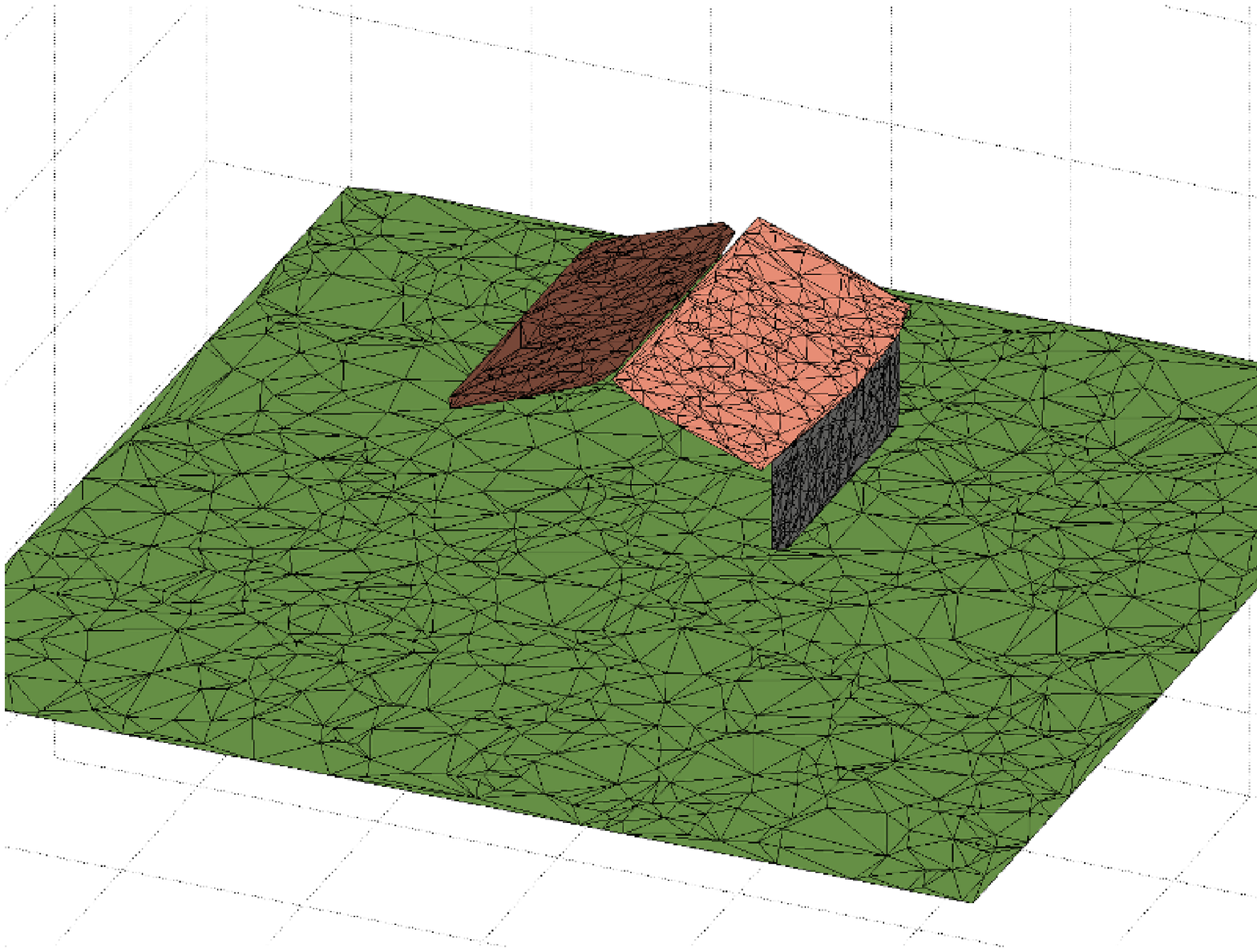} \\
      	
      	(a) Random planes & (b) Path & (c) Chessboard & (d) Indoors &(e) House
		\end{tabular}
	\caption{Images of the 5 synthetic datasets (upper row) and the resulting, extracted planar patches (bottom row).}
	\label{fig:Synthetic_results}
\end{figure*}

We have carried out a number of experiments in order to evaluate the robustness and applicability of our method for extraction of planar patches in 3D, with particular emphasis on robotic navigation, that is, extraction of the ground plane and segmentation of any approximately planar obstacles. First, we present the results from 5 synthetic datasets, where planar surfaces were arranged in different configurations of varying extraction complexity (see Fig. \ref{fig:Synthetic_results}). In all the synthetic cases, the ground truth (g.t.) plane equations and patch boundaries were available and used for accuracy comparison. 

Furthermore, in order to focus primarily on the extraction process we used a random subset from the available g.t. object vertices as corresponding, interest points in the two views, rather than using an interest operator and automatically establishing correspondences, something we did for the real datasets later on. We proceeded by adding Gaussian noise $\sigma$=0.001 in the pixel coordinates (image size 800$\times$600 pixels) and by using the triangulation approach of~\cite{HarStu1997} we obtained a noisy representation of the original, g.t. 3D cloud. We then used the segmentation method~\cite{ForMoe2006} to obtain the seed points and followed the procedure described in Alg. \ref{alg:algorithm}. The results are shown in Fig. \ref{fig:Synthetic_results}.

In all cases we have obtained well defined patches that closely adhere to the original segmentation but growth is not limited to the extracted elliptical boundaries. We can observe that there are very few outliers, spikes in the linear boundaries or encroachment regions between adjacent patches that share boundaries. In addition, no single point belongs to more than one patch. Note also that regions where no seed point was available (due to lack of stereo pair information) no patch has been initiated, but more encouraging, these points have not been assigned to any of the existing patches, something which would indicate a misclassification. 
The numerical errors (sum of squared differences SSD) between the g.t. plane coefficients and those recovered are shown in Table \ref{tab:synthetic_fit_errors}. The difference is very small and may be attributed to the effects of noise alone. 

\begin{figure}
 	\centering
	\includegraphics[width=0.23\textwidth]{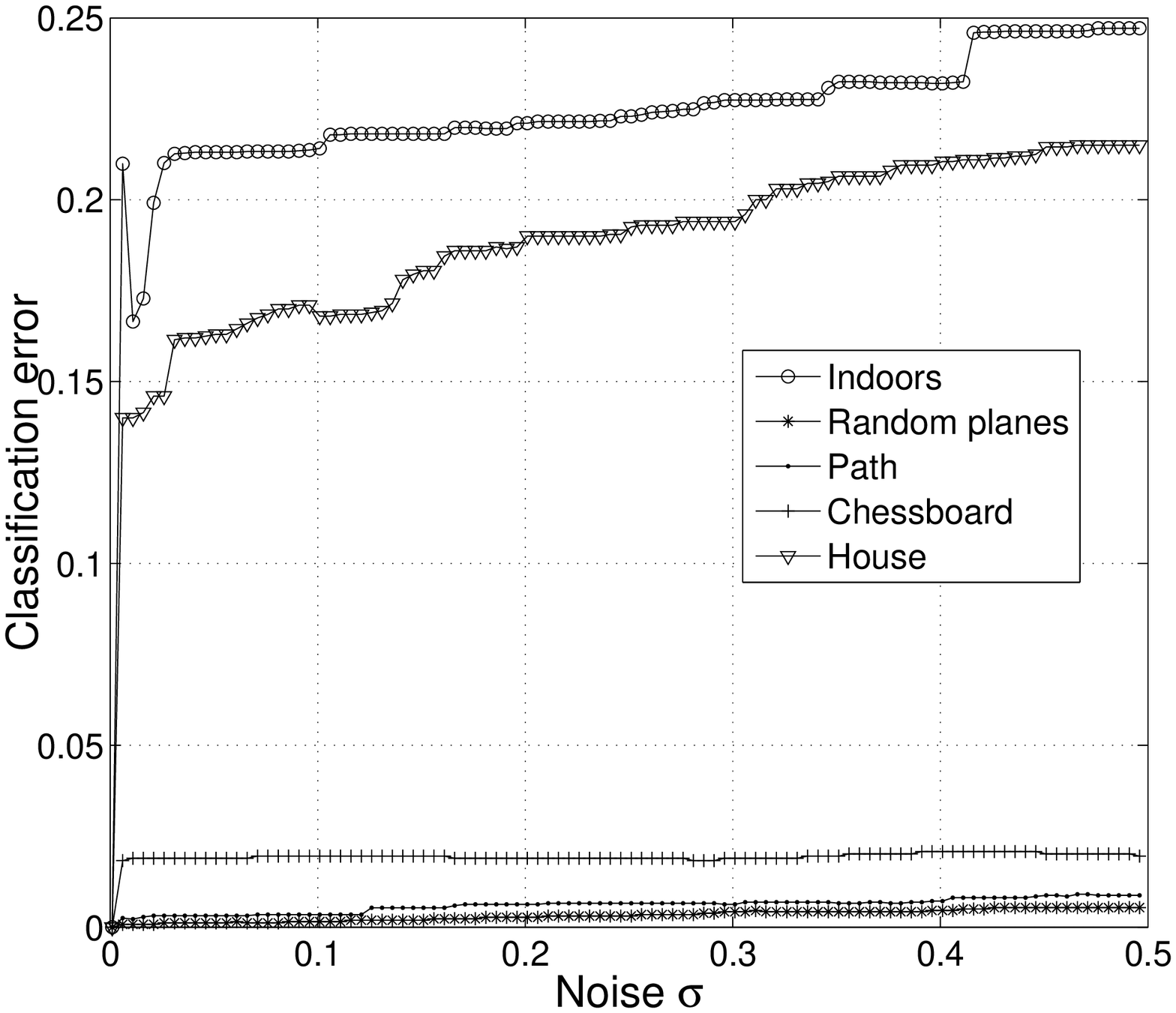} 
 	\includegraphics[width=0.23\textwidth]{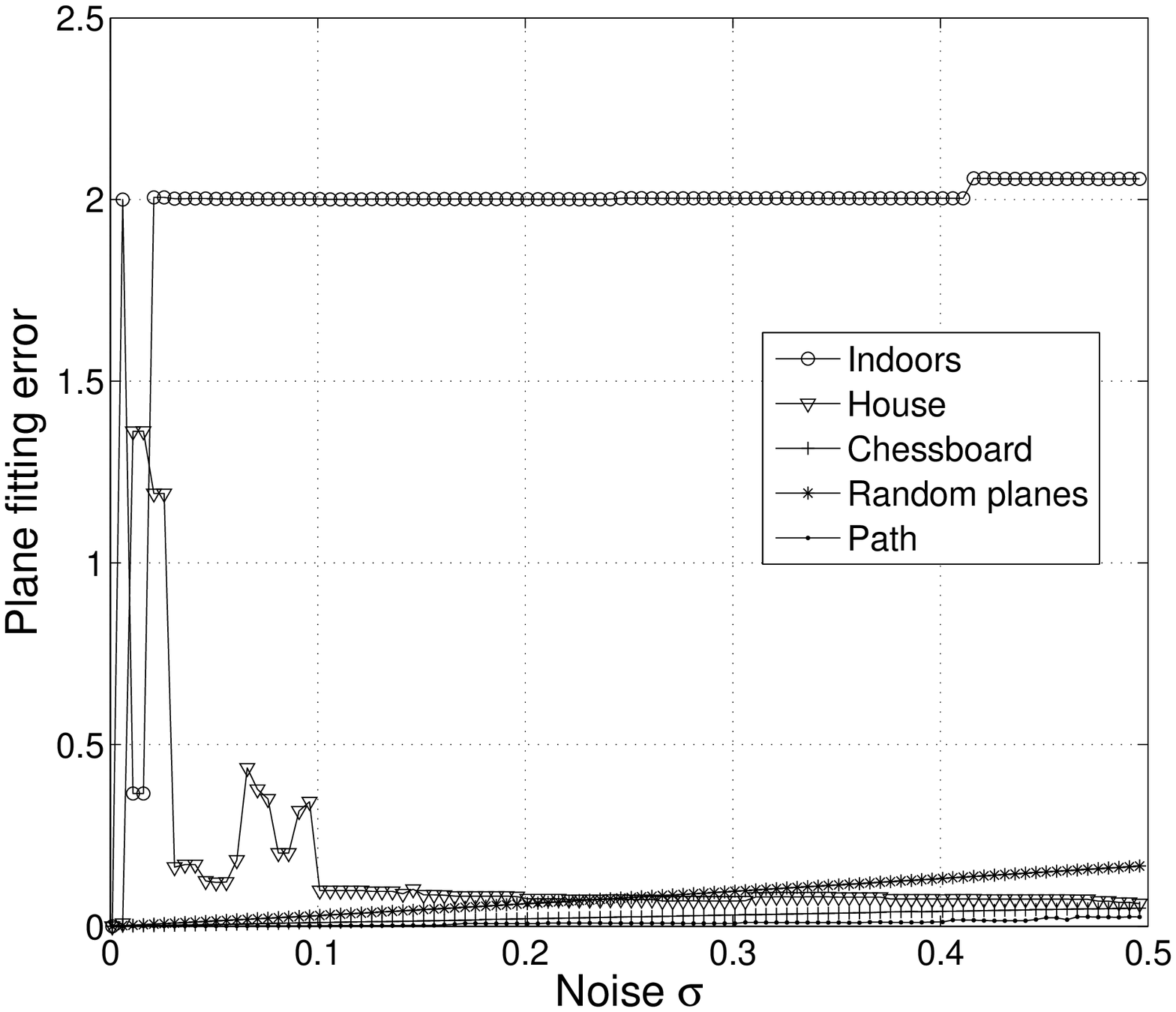}\\
	\caption{The classification and plane fitting robustness in relation to increasing noise.}
	\label{fig:Error_Plots}
\end{figure}

We have also tested the effects of increasing pixel noise on the extraction process, and particularly on the classification accuracy and SSD error in the plane coefficients. For all synthetic examples the thresholds remained fixed and tuned for $\sigma$=0.001, while we gradually increased the noise to 0.5 by taking 100 equidistant samples. The classification results for the 5 datasets are shown in Fig. \ref{fig:Error_Plots} (left). Here we define the error as 1-$n/N$ where $n$ are the points correctly classified for all classes and $N$ the total number of points. Any points that do not form part of any of the g.t. patches are considered to belong in a separate class. We see that generally the simpler datasets present easier extraction problems for our algorithm in the presence of noise with a near constant response. The plots for the more complex datasets although appear largely linear do have some spikes which indicate that an adjustment  of the thresholds is required, since the latter depends on the amount of noise.

The results are better when  we look at the SSD error in the extracted plane coefficients which are small and constant for all datasets (Fig. \ref{fig:Error_Plots} (right)). Once again the same two complex datasets show the familiar jump but only one of them maintains a high error. This graph is more indicative of the performance of our algorithm in the presence of noise and uncertainty, since by using the prior $P(d_c)$ from (\ref{eq:noise_prior}) we have favoured non-classification in order to maintain the correct plane equations.

\begin{figure*}
 	\centering
		\begin{tabular}{cccc}
				\includegraphics[width=0.2\textwidth]{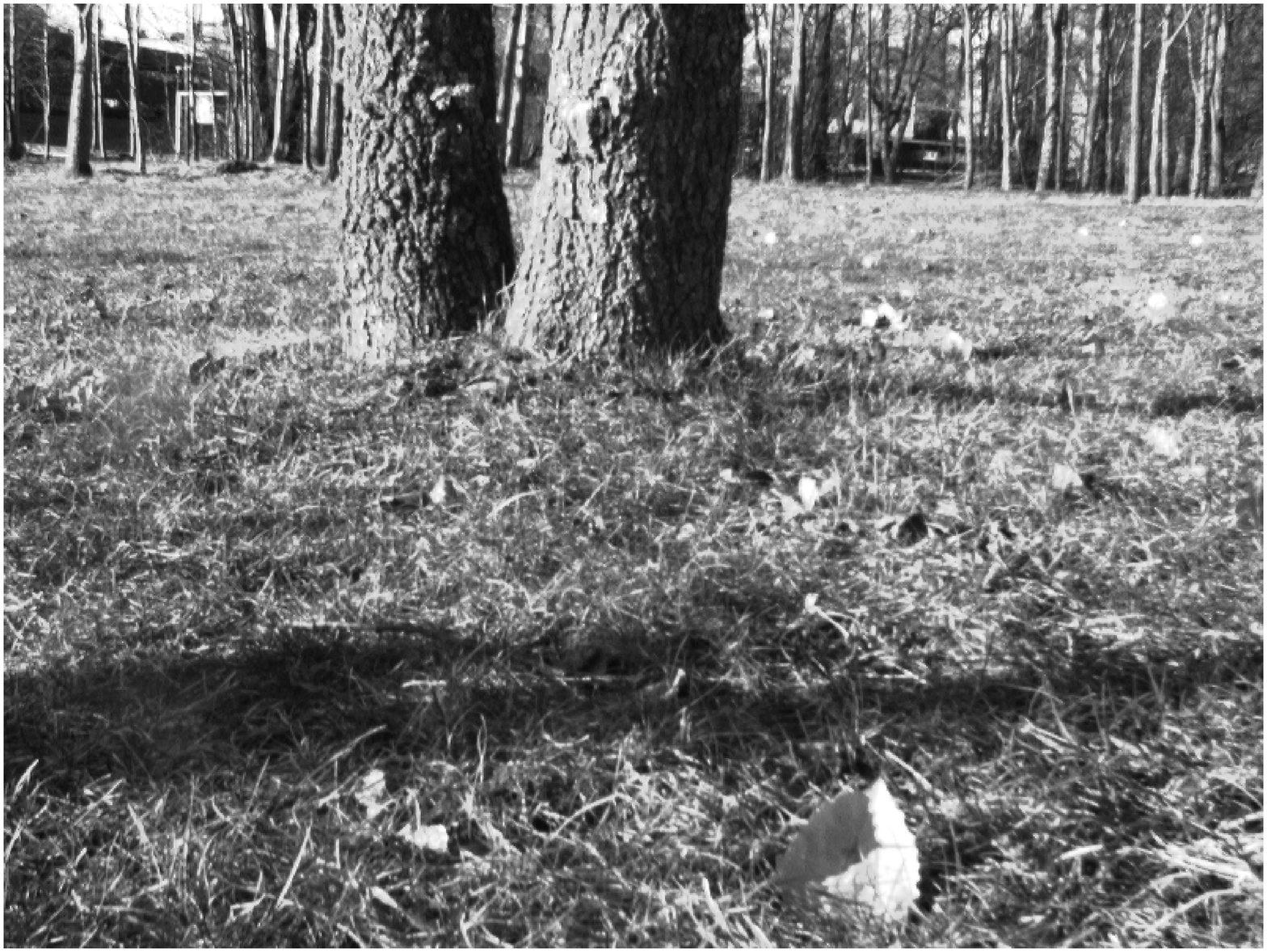} &
				\includegraphics[width=0.2\textwidth]{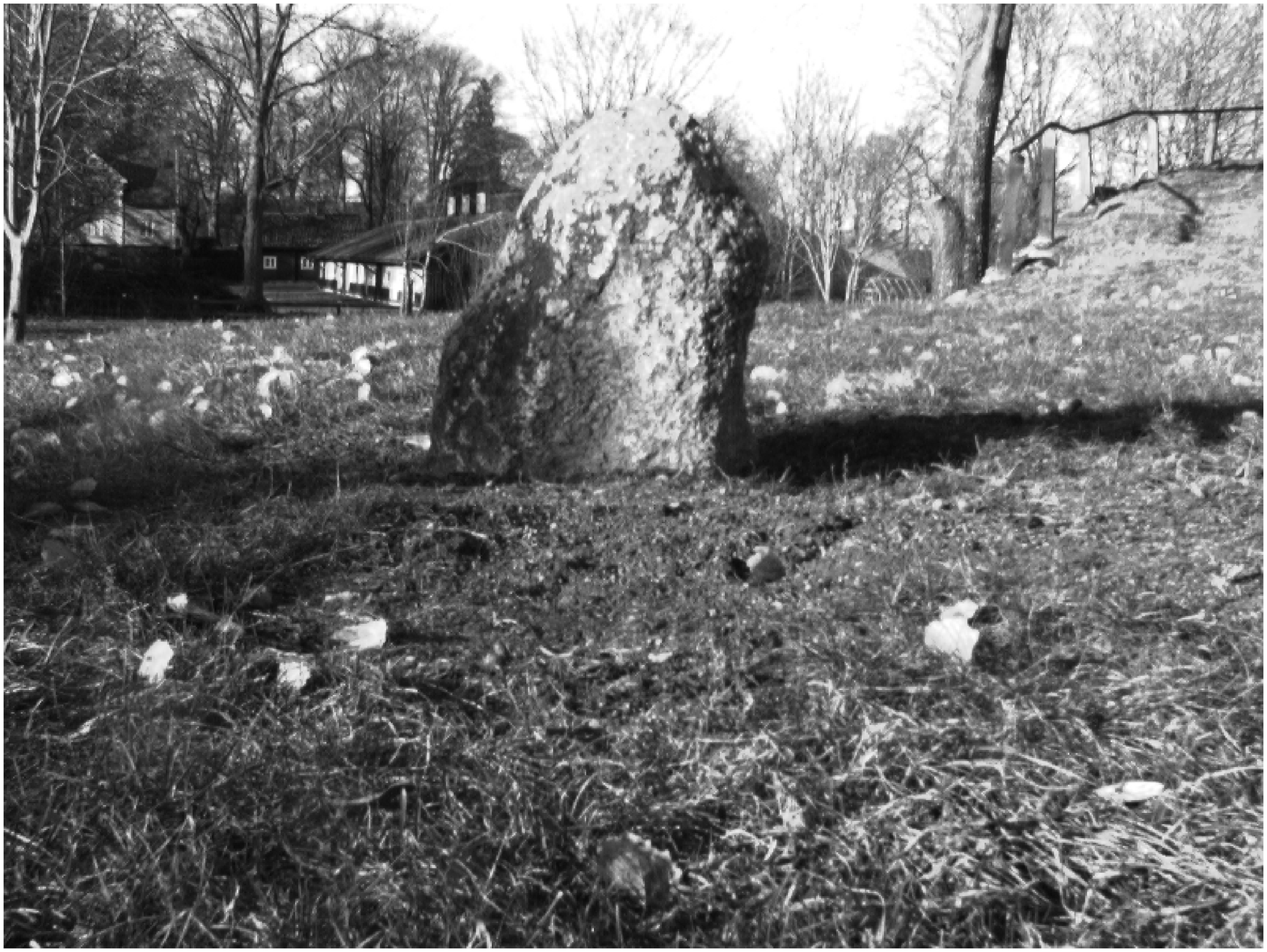} &
				\includegraphics[width=0.2\textwidth]{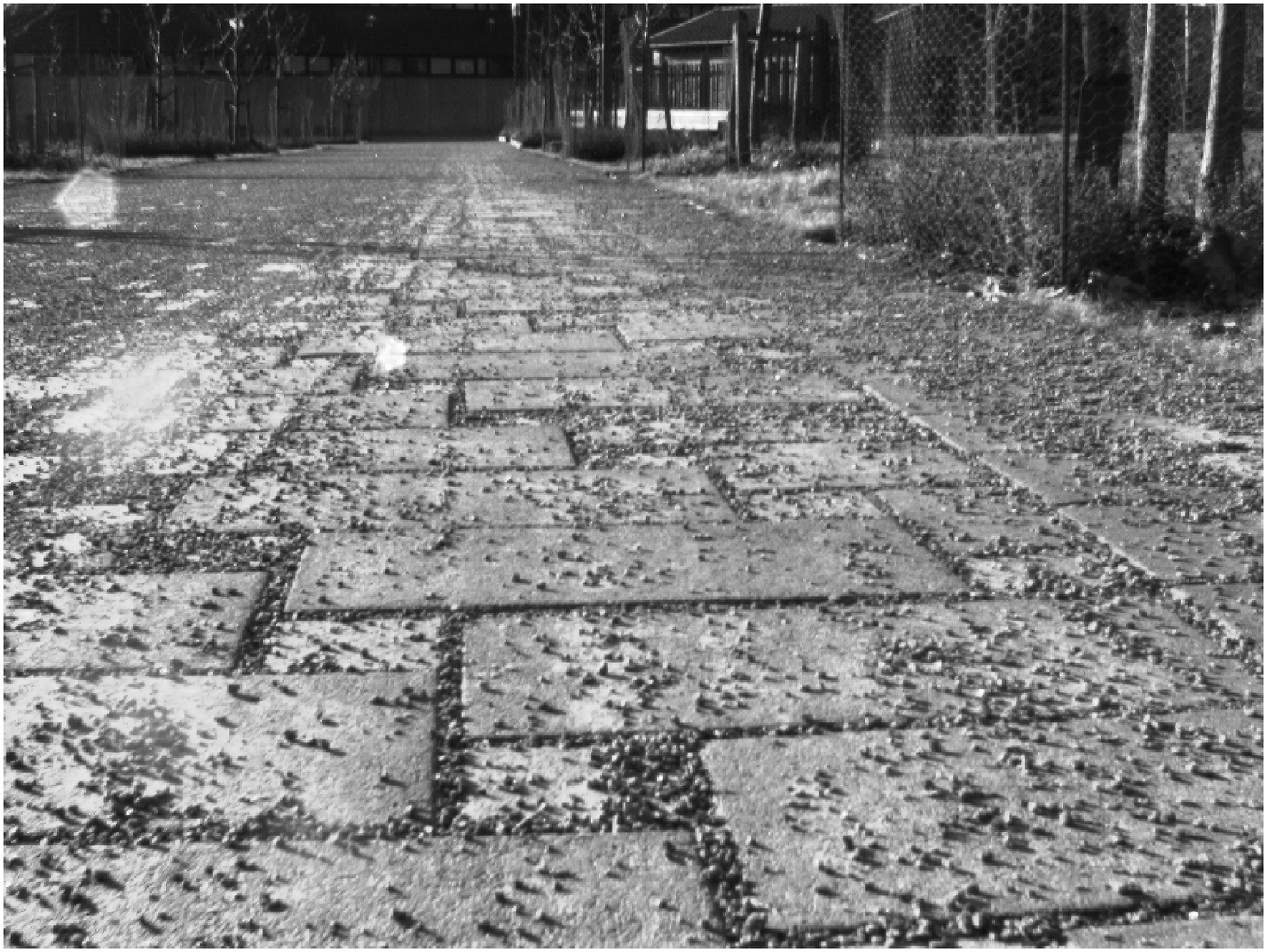} &
				\includegraphics[width=0.21\textwidth]{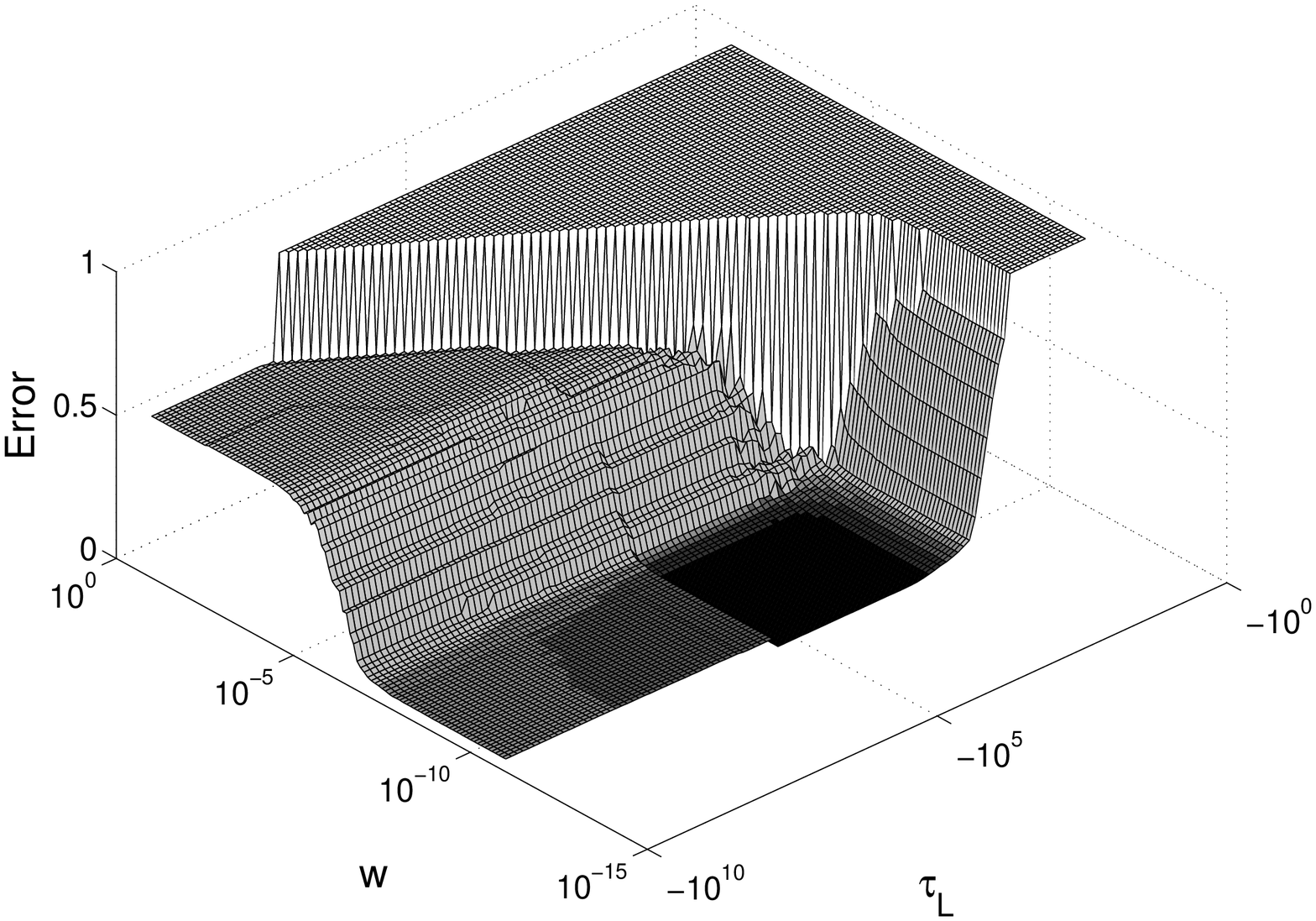} 
      	\\
      	\\
      	 &  &  & (a) Error vs choice of thresholds.\\
        \includegraphics[width=0.2\textwidth]{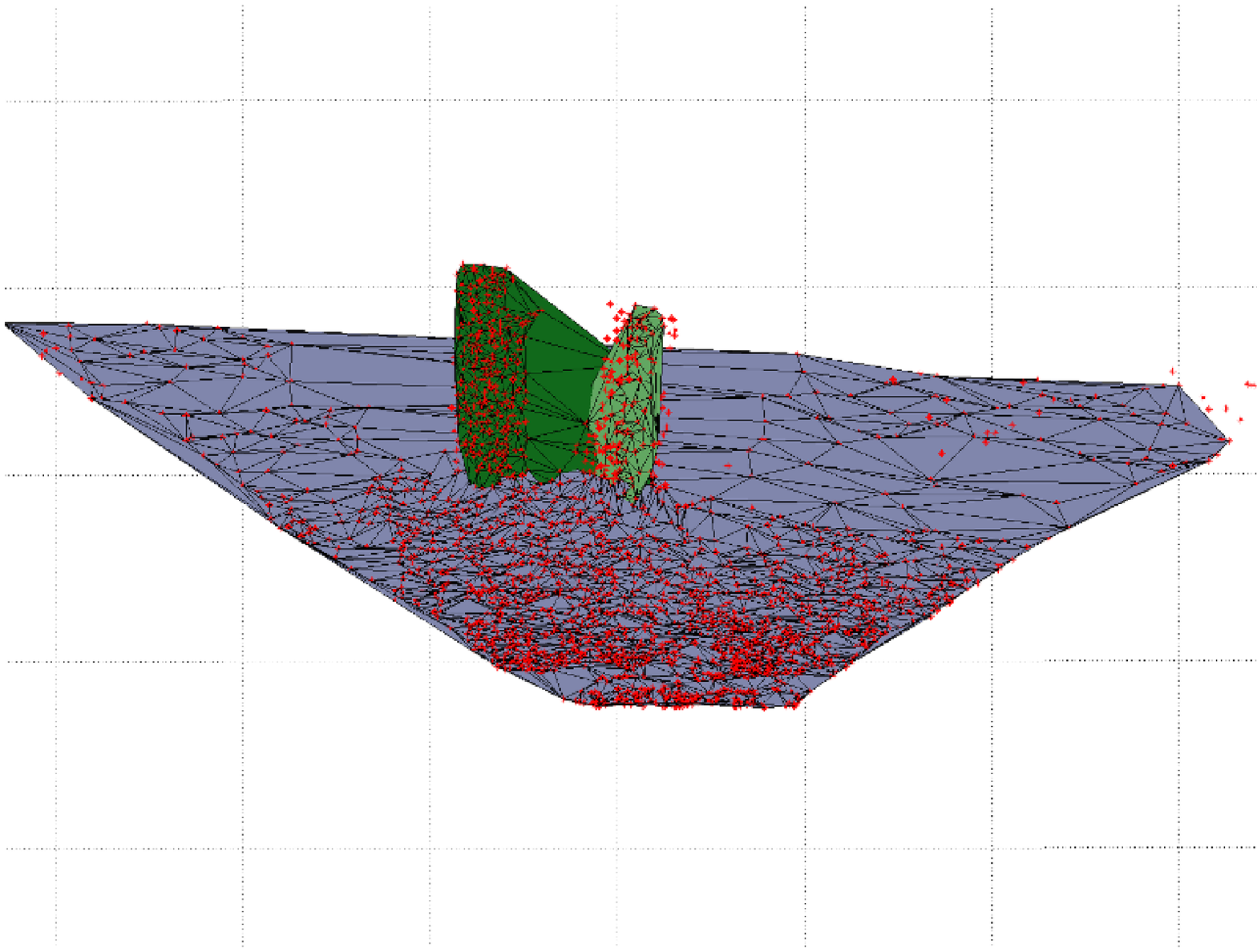} &
				\includegraphics[width=0.2\textwidth]{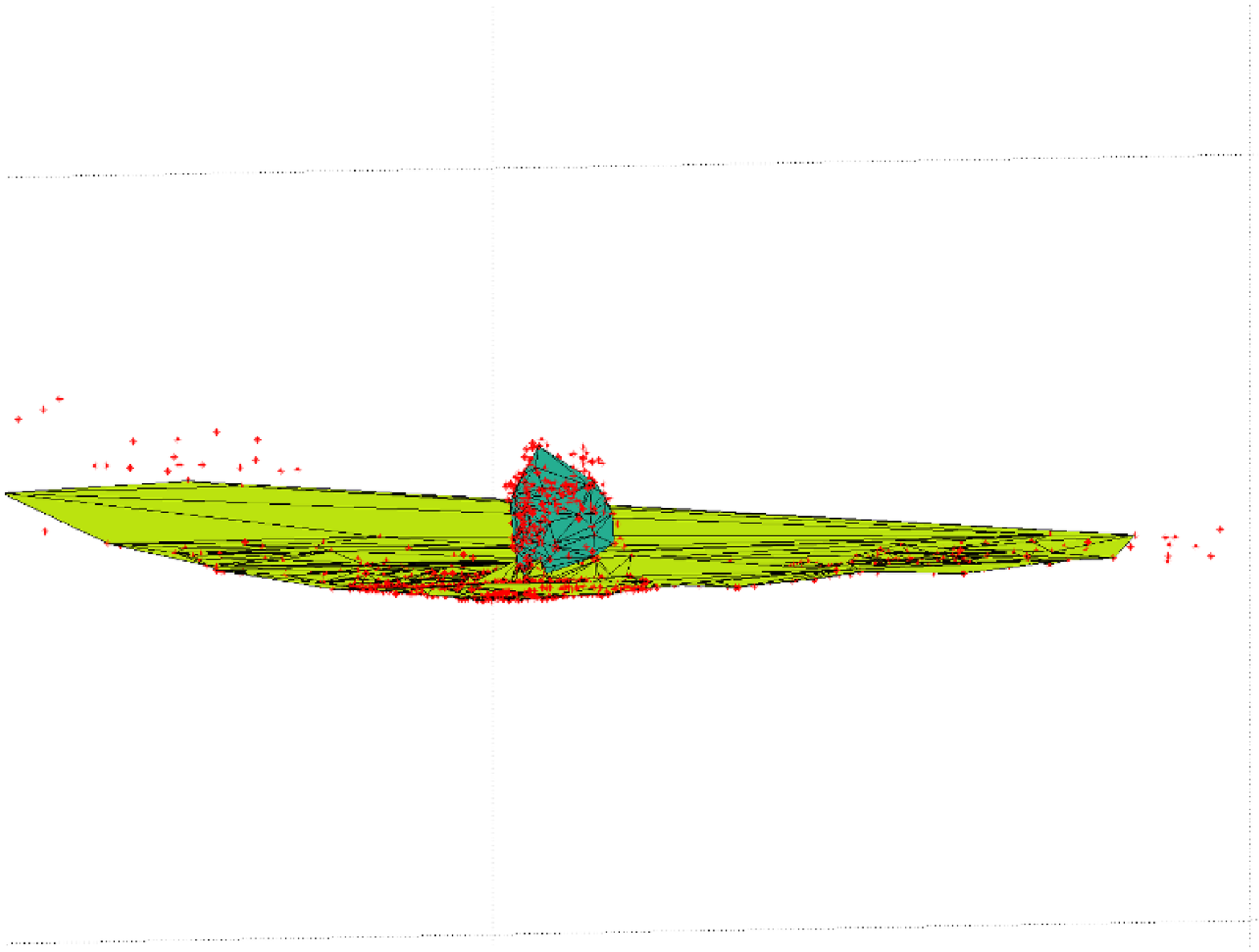} &
				\includegraphics[width=0.2\textwidth]{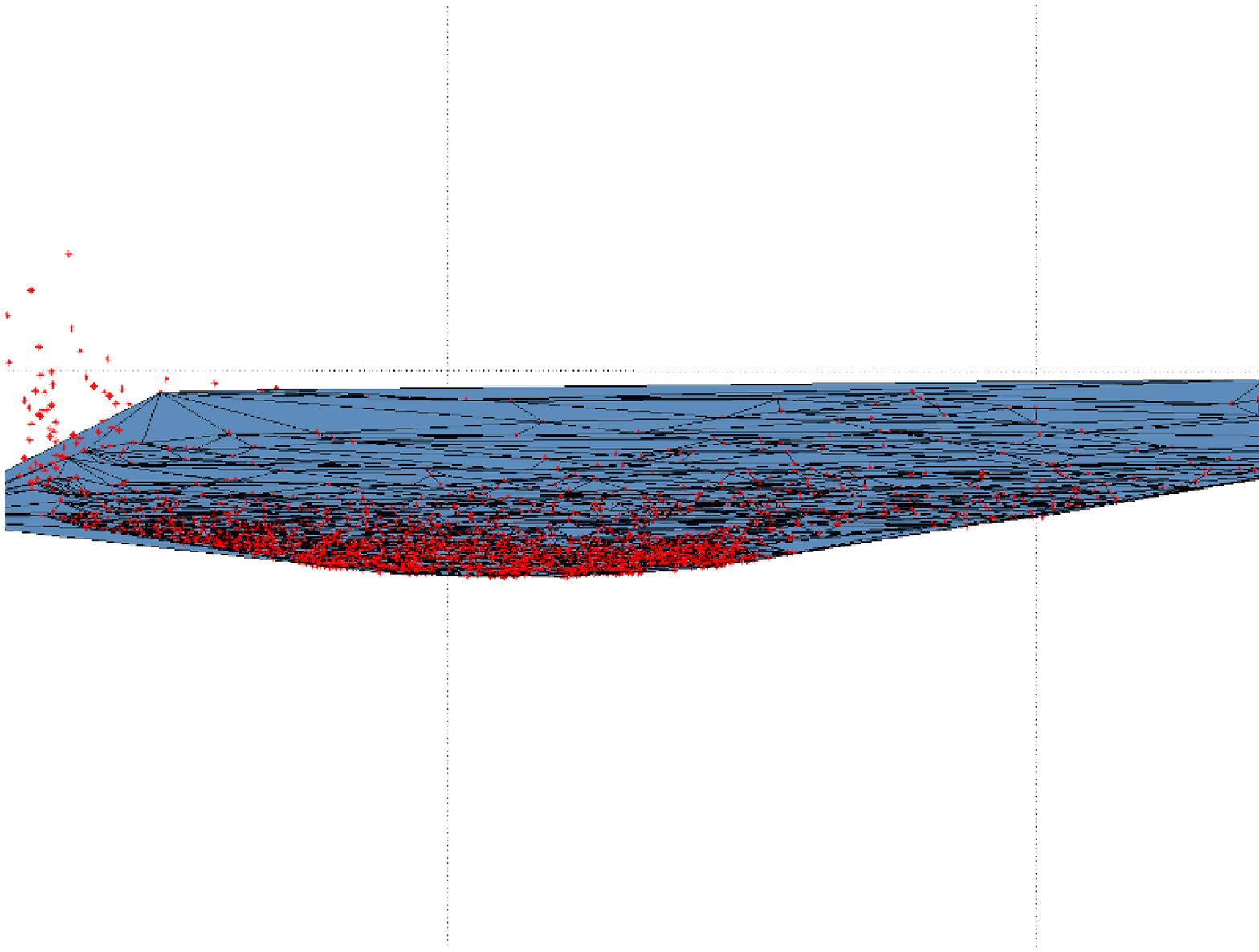}& 
				\includegraphics[width=0.21\textwidth]{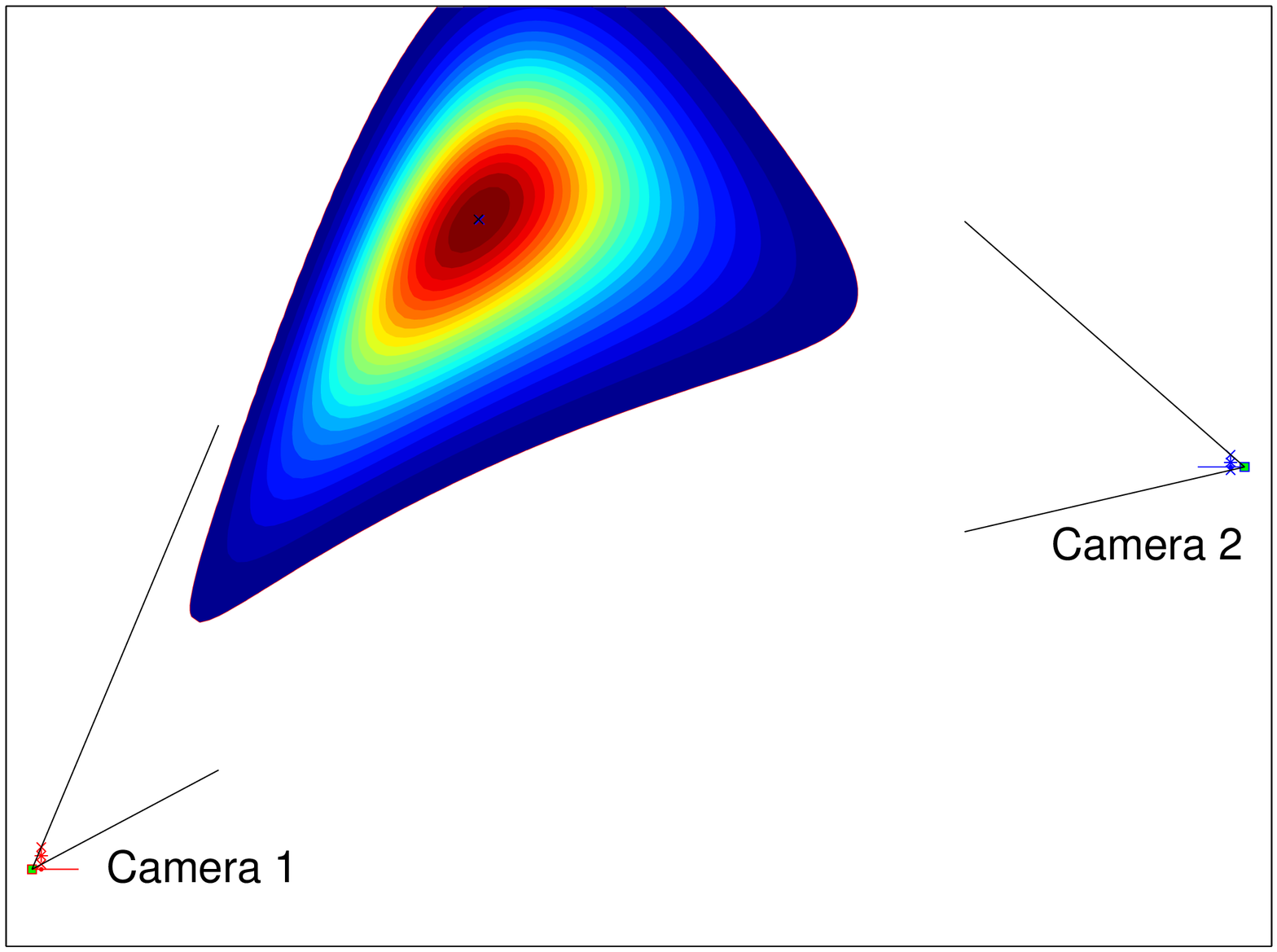} 
      	\\
      	 &  &  & (b) $P(\Upsilon|\bx,\bx')$ for Gaussian noise.
		\end{tabular}
	\caption{Real image data experiments (upper row) and the resulting, extracted planar patches (bottom row). Also  the error in relation to the thresholds $\tau_L$ and $w$ in log-space (a) and (b) the inherent uncertainty in triangulation with noise.}
	\label{fig:Real_results}
\end{figure*}

We also present a small number of results based on real data in Fig. \ref{fig:Real_results}. These are included so as to show that our method can perform well in more complicated, uncontrolled situations. In all these examples, we captured a number of forward motion sequences with a calibrated camera, which is a typical scenario of robotic navigation using monocular stereo. This is a difficult problem since the forward motion due to its reduced relative baseline, has a high inherent uncertainty for extraction of stereo information. We chose a number of Harris interest points~\cite{HarSte1998} and obtained frame correspondences using the KLT tracker~\cite{LucKan1981}. The points where triangulated and the patch extraction algorithm was used as before. 

The results are presented in Fig. \ref{fig:Real_results}. We can see that the dominant, ground plane is extracted successfully in all cases and at the correct position and orientation, despite the very noisy and inaccurate data. It is also the case that our algorithm appropriately penalises highly uncertain points in the distance in order to maintain overall accuracy of the plane. Furthermore, other obstacles such as the rock and trees are extracted with an approximate planar patch.

\begin{figure}
	\includegraphics[width=0.23\textwidth]{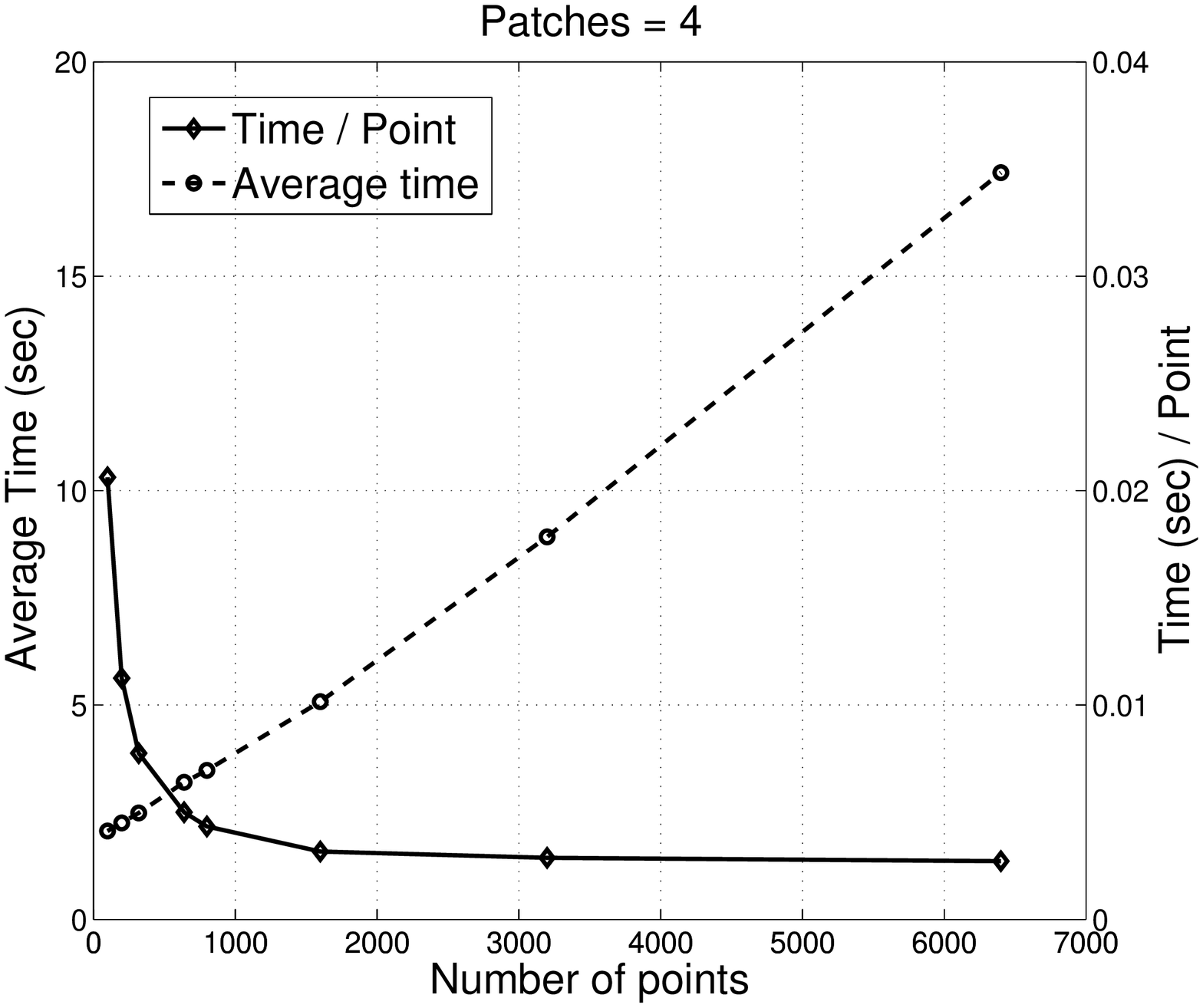} 
 	\includegraphics[width=0.23\textwidth]{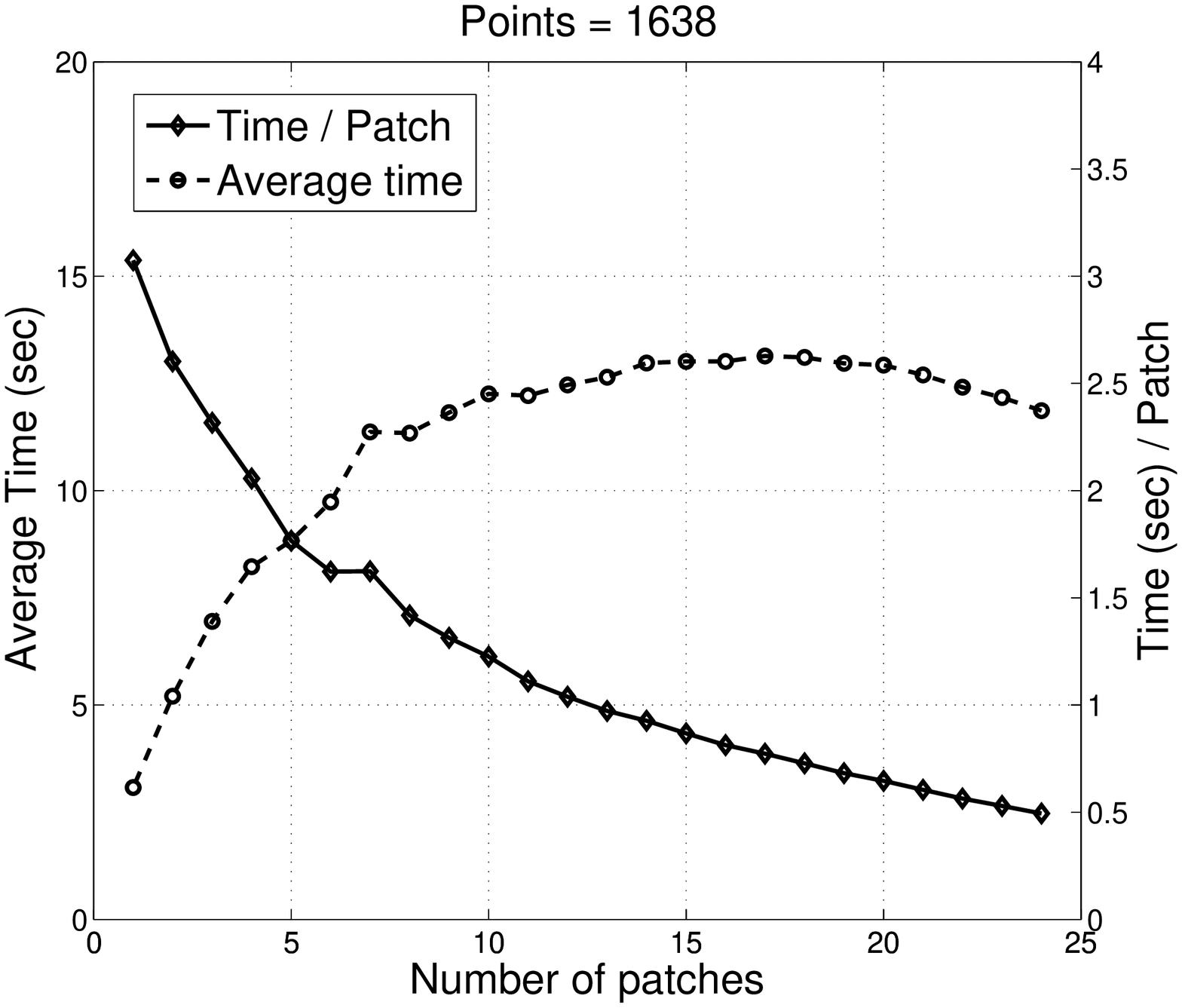}\\
 	\centering
	\caption{Typical performance results of our algorithm. Number of points vs. Time (left) and Number of patches vs. Time (right).}
	\label{fig:PointVSSpeed}
\end{figure}

Finally we carried out some speed tests to evaluate the overall efficiency of the method. In Fig. \ref{fig:PointVSSpeed} we see the execution time against an increasing number of extracted points for a fixed number of patches (left) and the time for an increased number of patches but fixed points (right). The implementation and speed testing were carried out on a   2.4Ghz CPU using Matlab.

\section{Conclusion and future work\label{sec:conclusion}}
We have presented a novel method for iteratively extracting planar patches from 3D data when (optional) stereo information is available. We begin with a 2D segmentation method based on average colour intensity, in order to obtain 3D seed points, with which to initialise our extraction algorithm, but also to approximate the boundaries of the patches. We then employ a Bayesian approach for patch growth based on the probability of the joint distance of a point from the plane and boundary edges of a patch. In addition, we use a penalty term based on the noise properties and settings of the stereo pair cameras in order to avoid classifying very uncertain points, thus maintaining overall patch quality.

We have experimented with a number of synthetic and real datasets with good results, while preserving the structural and perceptual integrity of the extracted regions. Furthermore, we have shown the effects of noise in the classification and plane fitting accuracy and robustness. Both are relatively robust for simplistic data, while more complicated datasets require adaptive threshold adjustment. We have carried out some initial investigation into the choice of appropriate thresholds and how these affect the patch extraction accuracy. The main difficulty is that these thresholds are interdependent and require a process of trial and error in order to be appropriately determined. Their optimal configuration occupies a small but relatively constant region of the log-space (see Fig. \ref{fig:Real_results} (a)) provided the noise in the data remains stable. This is a problem we would like to explore further in future work.

Furthermore, we would like to address the possibility of using additional characteristics for the patch segmentation and merging, such as texture, continuity, gradient and area in order to complement and enhance the current extraction method. We would also like to test other 2D segmentation approaches that are more robust to perspective and occlusion effects.

\begin{table}
\begin{centering}
\begin{tabular}{|l|c|c|}
\hline
{} & {\small Total error} & {\small Avg. plane error}\tabularnewline
\hline 
{\small Chessboard} & {\small 1.7E-7} & {\small 7.086E-9}\tabularnewline
\hline 
{\small House} & {\small 0.0067} & {\small 0.00168}\tabularnewline
\hline 
{\small Indoors} & {\small 0.00958} & {\small 0.00136}\tabularnewline
\hline 
{\small Path} & {\small 6.04E-9} & {\small 1.51E-9}\tabularnewline
\hline 
{\small Random planes} & {\small 4.76E-5} & {\small 6.8E-6}\tabularnewline
\hline
\end{tabular}\caption{SSD error between g.t. and synthetic tests.\label{tab:synthetic_fit_errors}}

\par\end{centering}

\end{table}

{\small
\bibliographystyle{ieee}
\bibliography{journals}
}

\end{document}